\documentclass[lettersize,journal]{IEEEtran}
\usepackage{cite}
\usepackage{amsmath,amssymb,amsfonts}
\usepackage{graphicx}
\usepackage{textcomp}
\usepackage{dsfont}
\usepackage{algpseudocode}
\usepackage{algorithm}
\usepackage{multirow, tabularx}
\usepackage{array}
\usepackage{subcaption}
\usepackage[caption=false,font=normalsize,labelfont=sf,textfont=sf]{subfig}

\begin{document}
\title{EEG\_GLT-Net: Optimising EEG Graphs for Real-time Motor Imagery Signals Classification}

\author{Htoo Wai Aung, Jiao Jiao Li, Yang An*, and Steven W. Su*, \IEEEmembership{Senior Member, IEEE} \thanks{Htoo Wai Aung is the first author of this paper, and he is with the School of Biomedical Engineering, Faculty of Engineering and IT, University of Technology Sydney, NSW 2007, Australia (e-mail: htoowai.aung@student.uts.edu.au).} 
\thanks{Jiao Jiao Li is with the School of Biomedical Engineering, Faculty of Engineering and IT, University of Technology Sydney, NSW 2007, Australia (e-mail: jiaojiao.li@uts.edu.au).}
\thanks{Yang An is with the Faculty of Engineering and IT, University of Technology Sydney, NSW 2007, Australia (e-mail: yang.an-1@student.uts.edu.au).}
\thanks{Steven W. Su is with the Faculty of Engineering and IT, University of Technology Sydney, NSW 2007, Australia ( e-mail: steven.su@uts.edu.au).}
\thanks{*Corresponding author.}
}

\markboth{Journal of \LaTeX\ Class Files,~Vol.~14, No.~8, August~2021}%
{Shell \MakeLowercase{\textit{et al.}}: A Sample Article Using IEEEtran.cls for IEEE Journals}

\maketitle

\begin{abstract}

Brain-Computer Interfaces (BCIs) connect the brain to external control devices, necessitating the accurate translation of brain signals such as from electroencephalography (EEG) into executable commands. Graph Neural Networks (GCN) have been increasingly applied for classifying EEG Motor Imagery (MI) signals, primarily because they incorporates the spatial relationships among EEG channels, resulting in improved accuracy over traditional convolutional methods. Recent advances by GCNs-Net in real-time EEG MI signal classification utilised Pearson Coefficient Correlation (PCC) for constructing adjacency matrices, yielding significant results on the PhysioNet dataset. Our paper introduces the EEG Graph Lottery Ticket (EEG\_GLT) algorithm, an innovative technique for constructing adjacency matrices for EEG channels. This method does not require pre-existing knowledge of inter-channel relationships, and it can be tailored to suit both individual subjects and GCN model architectures. We conducted an empirical study involving 20 subjects and six different GCN architectures to compare the performance of our EEG\_GLT adjacency matrix with both Geodesic and PCC adjacency matrices on the PhysioNet dataset. Our findings demonstrated that the PCC method outperformed the Geodesic approach by 9.65\% in mean accuracy, while our EEG\_GLT matrix consistently exceeded the performance of the PCC method by a mean accuracy of 13.39\%. Additionally, we found that the construction of the adjacency matrix significantly influenced accuracy, to a greater extent than GCN model configurations. A basic GCN configuration utilising our EEG\_GLT matrix exceeded the performance of even the most complex GCN setup with a PCC matrix in average accuracy. Our EEG\_GLT method also reduced Multiply-Accumulate Operations (MACs) by up to 97\% compared to the PCC method, while maintaining or enhancing accuracy. In conclusion, the EEG\_GLT algorithm marks a breakthrough in the development of optimal adjacency matrices, effectively boosting both computational accuracy and efficiency, making it well-suited for real-time classification of EEG MI signals that demand intensive computational resources.

\end{abstract}


\section{Introduction}
\label{sec:introduction}

\IEEEPARstart{B}{rain-Computer Interfaces} (BCIs) form an interdisciplinary bridge between engineering and neuroscience, enabling direct communication between the human brain and control devices. Originally designed to aid those with motor impairments \cite{WOLPAW2002767}, BCIs have expanded their applications to neurofeedback, gaming, and rehabilitation. Essentially, BCIs convert neural signals into actionable commands. The primary means of brain signal acquisition include electrocorticography (ECoG) and electroencephalography (EEG). Although ECoG boasts superior spatial resolution due to directly placing electrodes on the cortex, its invasive native limits its applications \cite{lebedev2006brain}. In contrast, EEG uses scalp placed electrodes to capture brain activity, making it more popular due to non-invasiveness and portability. This method captures various brain signals, from event-related to spontaneous and stimulus-evoked \cite{schomer2012niedermeyer}.

Motor Imagery (MI) pertains to the mental simulation of motor actions, such as moving one's hands or feet, without performing the actual movement \cite{hubbard2019eeg, mcfarland2000mu}. As highlighted by \cite{jeannerod1994representing}, action execution and its imagination share neural pathways. MI has prominent applications in rehabilitation and neuroscience. When paired with EEG, it captures neural signals generated from the intention to move. Integrating this with BCIs allows decoding EEG MI signals to control external devices such as a robotic exoskeleton. This technology is pivotal for those with motor impairments, especially stroke survivors, with the potential to restore quality of life and ability to perform daily activities. By accurately decoding EEG MI signals, BCIs can provide real-time feedback and communicate with assistive devices, to facilitate patient-intended movements \cite{biasiucci2018brain}.

Convolutional Neural Networks (CNNs) have consistently showcased superior results in computer vision tasks \cite{farabet2012learning, lecun1998gradient, lecun2010convolutional}. However, their effectiveness is largely constrained to regular Euclidean data, such as 2-dimensional grids and 1-dimensional sequences \cite{lecun2010convolutional}. A drop in capability is experienced with non-Euclidean data, primarily because CNN cannot accurately capture the intrinsic structure and connectivity of this data.

Graphs serve as powerful tools for representing relationships among entities, and are employed in diverse application areas including traffic systems, social networks, e-commerce platforms, biological structures, and trade networks. These graphs can highlight complex structures and be variable in nature such as being might be homogeneous or heterogeneous, having weight or not, and being signed or unsigned \cite{zhang2020deep}. The Graph Convolutional Neural Network (GCN) is an adaptation of CNN operations that is, tailored for graphs. GCN excel in managing non-Euclidean data, incorporating topological relationships during convolution. Their versatility allows application in a variety graph analysis tasks \cite{wu2020comprehensive, zhang2020deep}:
\begin{itemize}
    \item Node-Level Tasks: Involving prediction of nodes for either regression or classification.
    \item Edge-Level Tasks: Focusing on edge prediction, primarily for classification-tasks.
    \item Graph-Level Tasks: Targeting prediction of the entire graph for classification tasks.
\end{itemize}
Two main categories of GCNs are the spectral method \cite{bruna2013spectral, defferrard2016convolutional, levie2018cayleynets} and the spatial method \cite{hamilton2017inductive, monti2017geometric, niepert2016learning, gao2018large}. Studies \cite{bao2022linking} and \cite{shuman2013emerging} indicate challenges associated with the spatial method, particularly for matching local neighbourhoods. GCNs have an important application in classifying EEG signals at the graph level, where EEG readings from individual electrodes are treated as node attributes. With the help of GCNs, the inherent connections among electrodes can be integrated through the adjacency matrix, a capability beyond the reach of traditional CNNs.

EEG feature extraction is broadly categorised into time and frequency domain features. Building on the work of \cite{zeng2020hierarchy}, time-domain metrics such as Root Mean Square, skewness, minmax, variance, kurtosis, Hurst Exponent, Higuchi, and Petrosian fractal dimensions are derived within predefined time windows by \cite{meng2022electrical}. Within the frequency domain, emphasis is placed on power spectral density (PSD) and power ratio (PR) across specific frequency bands: $\delta$[0.5-4Hz], $\theta$[4-8Hz], $\alpha$[8-13Hz], $\beta$[13-30Hz], and $\gamma$[30-110Hz]. This is supplemented by other metrics such as total power, spectral entropy, and peak frequency, all captured within chosen time windows. In contrast to these window-based methods, Hou \textit{et al.} \cite{hou2022gcns} introduced a state-of-the-art time point classification approach in GCNs-Net by using each time point as features, offering a more time-resolved analysis for EEG-MI classification.

Establishing relationships between nodes is essential before deploying the GCN method. Studies \cite{zhang2022recognizing, jia2022efficient, wagh2020eeg} have utilised Geodesic distances between electrodes to form the adjacency matrix, while others \cite{bao2022linking, ma2023double, meng2022electrical, khaleghi2023developing, hou2022gcns} have employed the Pearson Coefficient Correlation (PCC) to assess correlations between EEG channels. Notably, \cite{song2018eeg} and \cite{bao2022linking} explored optimal adjacency matrices in EEG classification through a trainable matrix. Chen \textit{et al.} \cite{chen2021unified} introduced a unified GNN sparsification technique (UGS), giving rise to a Graph Lottery Ticket (GLT) by pruning both the original adjacency matrix and GNN weights. This method decreases the Multiply Accumulate (MAC) inference, thus reducing computational overhead, but with the problem of assuming that an initial adjacency matrix is provided. Our approach to classifying EEG MI signals, focuses on individual time points such as GCNs-Net \cite{hou2022gcns}. offering a granular and immediate signal analysis.

The primary contributions of this study can be summarised as:
 \begin{itemize}
    \item \textbf{EEG Graph Lottery Ticket (EEG\_GLT):} We present a novel method to construct an optimal adjacency matrix for EEG MI signal classification. Achieved through the iterative pruning of relationships among EEG channels, the EEG\_GLT introduces a new direction in EEG adjacency matrix design.
    
    \item \textbf{Channel Relationship Optimisation:} Our approach reveals the most advantageous relationship between EEG channels. It is tailored for catering to individual subjects and the architecture of GCN models, eliminating the need for prior knowledge about the inter-relationships among EEG channels.

    \item \textbf{Computational Efficiency:} Recognising the computational intensity of classifying EEG at single time points, our strategy mitigates the high demand for computational resources, proving especially beneficial for real-time applications.

    \item \textbf{Performance Validation:} We benchmark the accuracy of our EEG\_GLT method against two well-established techniques: the Geodesic method and the leading PCC method employed in the state-of-the-art GCNs-Net. This evaluation spans across six distinct spectral GCN models. Each model is distinguished by its unique specifications, including variations in GCN layer structures, polynomial degrees of filters, numbers of Fully Connected (FC) layers, and the amount of hidden nodes.
 \end{itemize}

\section{Methodology}
\subsection{Overview}

\begin{figure*}
    \centering
    \includegraphics[width=0.95\textwidth]{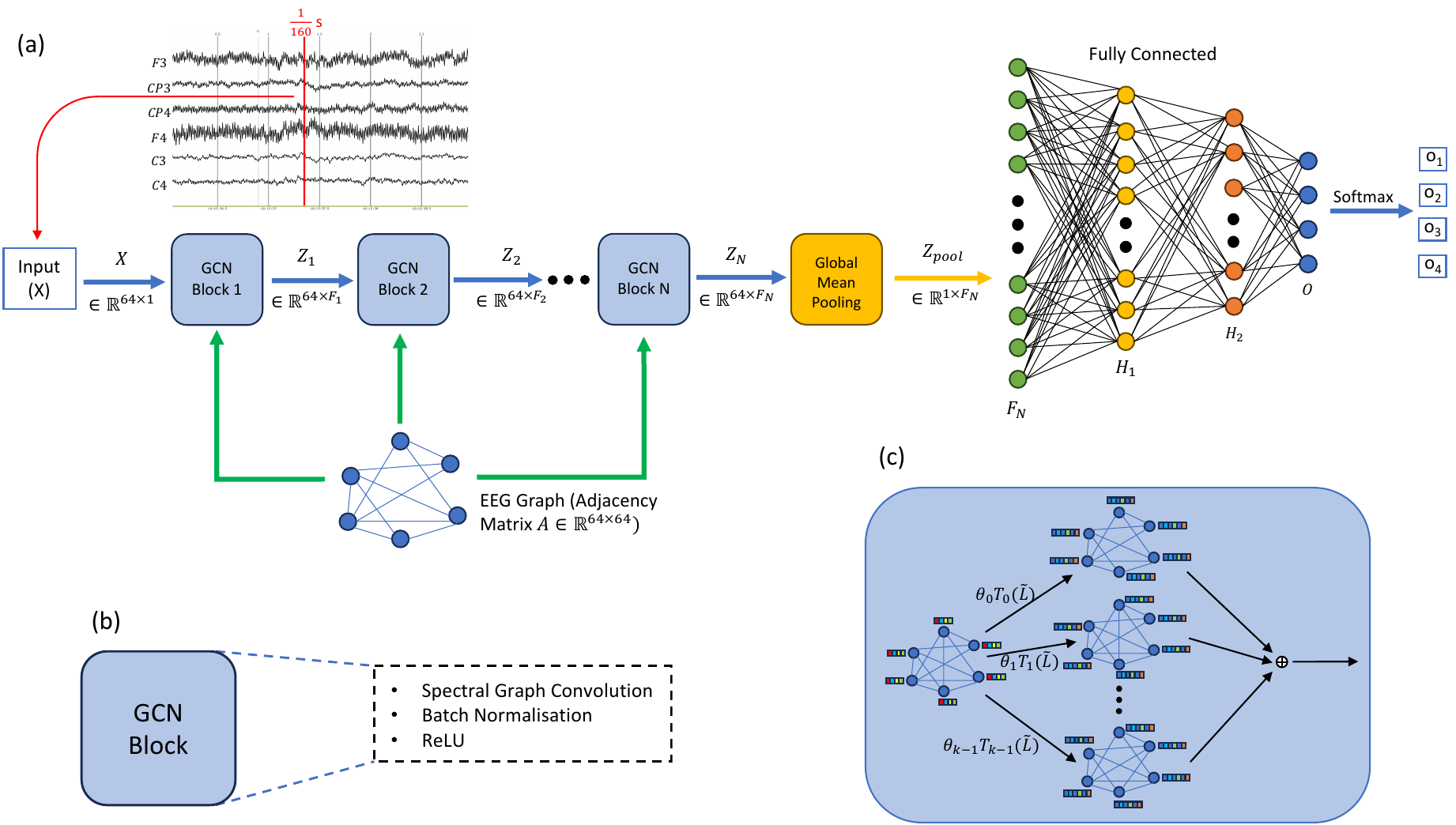}
    \caption{Our model: (a) Overall architecture (classifying EEG MI of one time point $\frac{1}{160}s$ of signals from 64 EEG electrodes). Note that EEG Graph adjacency matrix can be $A^{Geodesic}$, $A^{PCC}$ or $A^{EEG\_GLT}$, (b) Components inside the spectral graph convolution block, (c) Chebyshev spectral graph convolution}
    \label{fig: overall_archi}
\end{figure*}

As shown in Figure~\ref{fig: overall_archi}, the project framework was as follows:
\begin{itemize}
    \item EEG signals from 64 channels were captured at each time point $\frac{1}{160}s$ and used as input features for the EEG\_GLT-Net.
    \item Additionally, the EEG\_GLT-Net accepted the graph representation as another form of input. This representation included the graph Laplacian, derived using three different methods: Pearson Correlation Coefficient (PCC) between EEG channels, Geodesic distance between EEG electrodes, and our newly proposed EEG Graph Lottery Ticket Adjacency Matrix Mask $(m_{EEG\_GLT})$.
    \item The EEG\_GLT-Net processed these inputs to decode the EEG MI time point signal, which was then categorised into one of the four MI types.
\end{itemize}

\subsection{Dataset Description}
This paper utilised the PhysioNet EEG Motor Imagery (MI) dataset \cite{physionet_dataset} encompassing over 1,500 EEG recordings sourced from 109 participants. The recordings were captured using 64 EEG electrodes, consistent with the international 10-10 system, with the exclusion of F9, Nz, F10, FT9, FT10, A1, A2, TP9, TP10, P9, and P10 channels. Each participant executed 84 trials, broken down into 3 runs with, 7 trials per run, spanning 4 distinct tasks. The tasks included:
\begin{itemize}
    \item Task 1: Imagining the act of opening and closing the left fist.
    \item Task 2: Imagining the act of opening and closing the right fist.
    \item Task 3: Imagining the act of opening and closing both fists simultaneously.
    \item Task 4: Imagining the act of opening and closing both feet.
\end{itemize}
Recordings in the dataset were originally sampled at 160 Hz and each recording had a duration of 4 seconds. Our study employed time point samples for classification, and our analysis was strictly conducted at the subject level. Although the original dataset comprised 109 participants, our study focused solely on 20 subjects, labelled $S_{1}$ to $S_{20}$.

\subsection{Data Pre-processing and Feature Extraction}
In the initial pre-processing phase, raw signals underwent only a notch filter at the 50Hz power line frequency, foregoing typical filtering or denoising steps to maximise data integrity. Although each task lasted for a 4-second duration, only the time period from $t=1s$ to $t=3s$ was considered in our experiments. This is because subjects typically exhibited greater readiness post $t=1s$. All 64 EEG channels were incorporated into our model. We utilised the signal values from each EEG channel at each time point as features for individual nodes. The construction methods of the adjacency matrix, which captures brain connectivity, are elaborated in Sections~\ref{subsubsec: graph_representaion} and \ref{subsec: EEGGLT}. The training data underwent normalisation, ensuring a mean $\mu=0$ and a standard deviation $\sigma=1$ for each channel. Following this, both the test and validation sets were adjusted in alignment with the normalisation parameters established from the training data.

\subsection{Graph Preliminary}\label{subsubsec: graph_representaion}
\subsubsection{Graph Representation}
Consider a directed weighted graph represented as $G=\{V,E\}$. Here, $|V|=N$ denoted the number of nodes and $|E|$ was the count of edges connecting the nodes. The node set was defined as $V=\{v_1,v_2,…,v_n\}$ and the node feature matrix of the entire graph was represented by $X\in\mathbb{R}^{N\times F}$. The adjacency matrix, denoted as $A\in\mathbb{R}^{N\times N}$, captured the graph’s overall topology. Specifically, if an edge existed between nodes $v_i$ and $v_j$ (i.e., $(v_i,v_j)\in E$), then $A[i,j]\neq0$. Otherwise, $A[i,j]=0$.

The adjacency matrix for the PCC method was defined in Equation~\ref{eqn: pcc adj}, where $I$ was the identity matrix and $|P|$ was the absolute PCC matrix as in Equation~\ref{eqn: pcc P}. This PCC matrix, $|P|\in[0,1]$, captured the linear correlations among EEG channel signals.

\begin{equation}
\label{eqn: pcc P}
P_{ij}=\frac{cov(x_i, x_j)}{\sigma_i \sigma_j}
\end{equation}

\begin{equation}
\label{eqn: pcc adj}
A^{PCC} = |P|-I
\end{equation}

For the Geodesic-distance adjacency matrix method, the configuration of 64 electrodes into a unit sphere acted as a stand-in for spatial brain connectivity. This allowed the computation of geodesic distances between the electrodes placed on a sphere of radius $r$. If two electrodes have Cartesian coordinates $(x_i,\ y_i,\ z_i)$ and $(x_j,\ y_j,\ z_j)$, the geodesic distance for the adjacency matrix was calculated using Equation~\ref{eqn: geodesic Adj}. These distances were standardised into the [0, 1] range.

\begin{equation}
\label{eqn: geodesic Adj}
A_{ij}^{Geodesic}=arcos(\frac{(x_i\ x_j+y_i\ y_j+z_i\ z_j)}{r^2})
\end{equation}

The degree matrix, $D$, was a diagonal representation of $A$, where the $i^{th}$ diagonal element of $D$ was computed as $D_{ii}=\sum_{j=1}^{N}A_{ij}$. The combinatorial Laplacian matrix, $L\in\mathbb{R}^{N\times N}$, was described as $L=D-A$. A normalised version of this combinatorial Laplacian can be obtained using:

\begin{equation}
\label{eqn: laplacian norm}
L=I_{N} - D^{-1/2} A D^{-1/2}
\end{equation}

\subsubsection{Spectral Graph Filtering}
The eigenvectors of the graph Laplacian matrix can be expressed as graph Fourier modes, with $\{u_l\}_{l=0}^{N-1}\in \mathbb{R}$. The diagonal matrix of these Fourier frequencies, $\Lambda$, is given by $diag[\lambda_0 , ..., \lambda_{N-1}]\in \mathbb{R}^{N\times N}$. We defined the Fourier basis, $U=[u_0, ..., u_{N-1}]\in \mathbb{R}^{N\times N}$, which allows for the decomposition of the Laplacian matrix, $L$, into $L=U \Lambda U^{T}$. The signal $x$ can be transformed by graph Fourier into $\hat{x} \in \mathbb{R}^{N} $ using $\hat{x}=U^{T}x$, while the inverse graph Fourier transform is given by $x=U\hat{x}$. The convolution operation on graph $G$ is defined as:

\begin{equation}
\label{eqn: graph conv 1}
x *_{G} g = U((U^T x) \odot (U^T g))
\end{equation}

\noindent where $g$ represents the convolutional filter and $\odot$ denotes the Hadamard product. Given that $g_\theta (\Lambda)=diag(\theta)$, where $\theta\in \mathbb{R}^N$ represents the vector of Fourier coefficients, the Graph convolution operation can be implemented as follows:

\begin{equation}
\label{eqn: graph conv 2}
x *_G g_\theta = U g_\theta (\Lambda) U^T x
\end{equation}

\noindent where $g_\theta$ is a non-parametric filter, and polynomial approximation is employed to mitigate the excessive computational complexity. Chebyshev graph convolution, a specific instance of graph convolution, utilises Chebyshev polynomials for filter approximation, thereby reducing computational complexity from $O(N^2)$ to $O(KN)$ \cite{defferrard2016convolutional}. The approximation of $g_\theta (\Lambda)$ under the $K^{th}$ order Chebyshev polynomial framework is given by:

\begin{equation}
\label{eqn: graph cheby}
g_\theta (\Lambda) =  \sum_{k=0}^{K-1}\theta_k T_k (\hat{\Lambda})
\end{equation}

\begin{equation}
\label{eqn: A_lambda norm}
\hat{\Lambda}=\frac{2\Lambda}{\Lambda_{max}} - I_N
\end{equation}

Normalising $\Lambda$ can be achieved by using Equation~\ref{eqn: A_lambda norm}, where $\Lambda_{max}$ denotes the largest entry in the diagonal of $\Lambda$, and $I_N$ represents the diagonal matrix of the scaled eigenvalues. In the equation above, $\theta_k$ refers to the Chebyshev polynomial's coefficients, and $T_k(\hat{\Lambda})$ is obtained by the following equations:

\begin{equation}
\label{eqn: graph cheby norm}
\{T_0(\hat{\Lambda})=1,  T_1=(\hat{\Lambda}), T_k(\hat{\Lambda})=2\hat{\Lambda}T_{k-1}(\hat{\Lambda})-T_{k-2}(\hat{\Lambda})\}
\end{equation}

Finally, the signal $x$ can be convolved with the defined filter $g_\theta$ as follows:

\begin{equation}
\label{eqn: graph cheby 2}
\begin{split}
x *_G g_\theta = U \sum^{K-1}_{k=0}\theta_k T_k(\hat{\Lambda})U^T x \\
x *_G g_\theta=\sum^{K-1}_{k=0} \theta_k T_k (\widetilde{L})x
\end{split}
\end{equation}

The normalised Laplacian matrix, denoted as $\widetilde{L}$ can be computed using the following Equation~\ref{eqn: lambda norm}.

\begin{equation}
\label{eqn: lambda norm}
\widetilde{L}=\frac{2L}{\lambda_{max}}-I_N
\end{equation}

\subsection{EEG Graph Lottery Ticket (EEG\_{GLT})}\label{subsec: EEGGLT}
In the process of executing a forward pass with the spectral GNN function, symbolised as $f(., \Theta)$, and given a graph denoted as $G=\{A, X\}$, the method presented in UGS method \cite{chen2021unified} aims to search the adjacency matrix mask $m_g\in\{0, 1\}$ with the maximum sparsity that concurrently maintained the highest prediction accuracy. In our model, the original matrix $A_{original\_ij}= \{0, if \ \ i=j; 1, otherwise\}$ in the shape of $|V|\times|V|$ was not trainable. The adjacency matrix mask in our model $m_g\in\mathbb{R^{|V|\times|V|}}$ was trainable.

\begin{equation}
\label{eqn: adj_matrix}
A = A_{original}\odot m_g
\end{equation}

\begin{figure}
    \centering
    \includegraphics[width=0.48\textwidth]{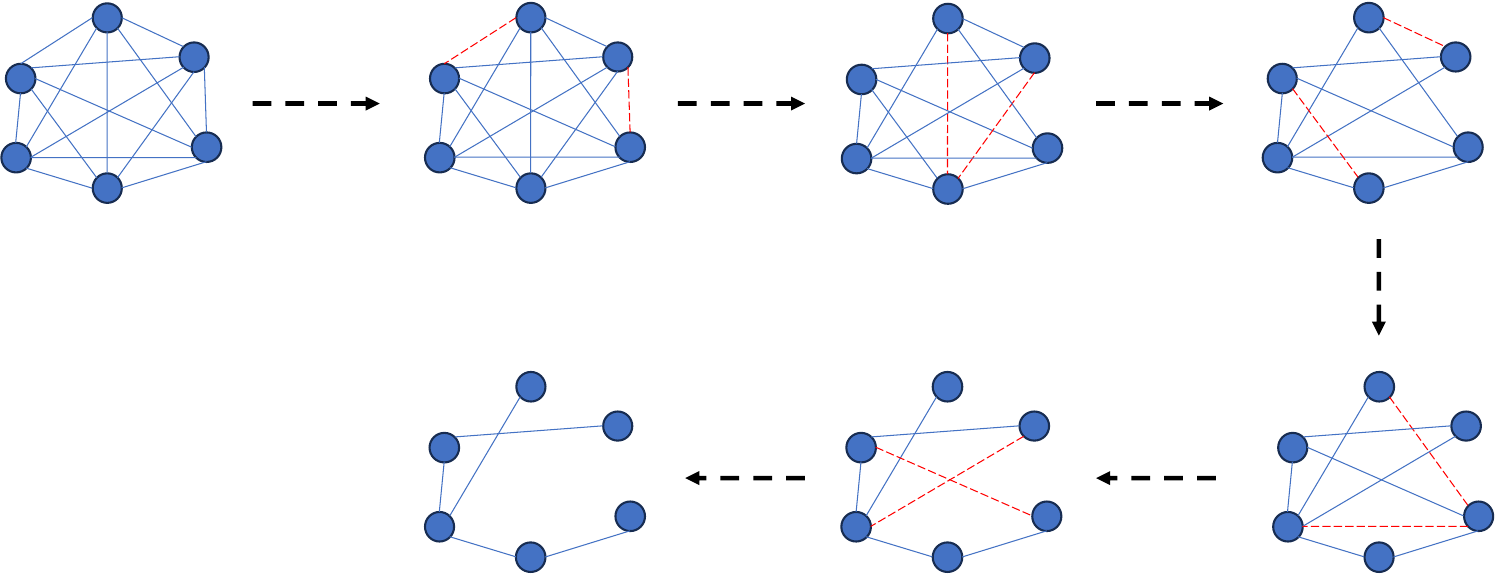}
    \caption{EEG graph ($m_g$) pruning using Algorithm 1: At each $N_{ep}$ iteration, the bottom $p_g$\% are pruned, reducing density from 100\% until the lowest density $s_g$\%. Solid lines indicate remaining edges, while red-dashed lines depict removed edges}
    \label{fig: graph_prune}
\end{figure}

Once the model had undergone $N$ epochs, the lowest $p_g$\% $(p_g=10\%)$ of the values in the trained $m_g$ at highest accuracy of the validation dataset were pruned. These values were set to 0, while the remaining values were set to 1 as shown in Figure~\ref{fig: graph_prune}. Concurrently, the spectral filter weights, represented as $\Theta$, were reset to their initial state, $\Theta_0$. The trained $m_g$ that yielded the highest accuracy of the validation set within the span of $N$ epochs was designated as the Graph Lottery Ticket (GLT) and duly noted. This process continued, and a GLT was recorded for each level of graph sparsity until the sparsity of $m_g$ fell below the pre-determined final sparsity level, $s_g$. The EEG\_GLT was ultimately identified as the GLT that achieves the highest accuracy alongside the highest level of graph sparsity. Moreover, it delineated the optimal adjacency matrix capable of producing the highest accuracy.

\begin{algorithm}[!ht]
\caption{Finding EEG Graph Lottery Ticket}
\label{alg: EEG_GLT}
\noindent \textbf{Input:} \parbox[t]{\dimexpr\linewidth-\algorithmicindent}{Graph $G=\{A, X\}$, GNN $f(G,\Theta)$, GNN initialisation \newline $\Theta_{0}$, $A_{original\_ij}= \{0, if \ \ i=j; 1, otherwise\}$, \newline initial Adjacency Matrix Mask $m_{g}^{0}=A_{original}$, \newline learning rate $\eta=0.01$, pruning rate $p_{g}=10 \%$, \newline pre-defined lowest Graph Density Level $s_{g}=13.39\%$. \strut}
\noindent \textbf{Output:} \parbox[t]{\dimexpr\linewidth-\algorithmicindent}{EEG Graph Lottery Ticket $(m_{g\_EEG\_GLT}) - m_{g}^{s,i}$ \newline at the highest accuracy with the highest sparsity \newline possible.\strut}
\begin{algorithmic}[1]
\While {$\frac{||m_g^s||_0}{||A_{original}||_0}\geq s_g$}
\For{for iteration $i=0, 1, 2, ..., N_{ep}$}
\State \parbox[t]{\dimexpr\linewidth-\algorithmicindent}{Forward $f(.,\Theta_{i})$ with $G_{s}=\{m_{g}^{s,i}\odot A_{original}, X\}$ \newline to compute Cross-Entropy Loss, $L$ \strut}

\State Backpropagate and update, $\Theta_i$ and $m_{g}^{s,i}$ using Adam Optimizer
\EndFor

\State \parbox[t]{\dimexpr\linewidth-\algorithmicindent} {Record $m_{g}^{s,i}$ with the highest accuracy in validation set during the $N_{ep}$ iteration \strut}
\State \parbox[t]{\dimexpr\linewidth-\algorithmicindent}{Set $p_{g}=10 \%$ of the lowest absolute magnitude values in $m_{g}^{s}$ to 0 and the others to 1, then obtain a new $m_{g}^{s+1,0}$ \strut}

\EndWhile

\end{algorithmic}
\end{algorithm}

\subsection{General Model Architecture}
A GCN structure was designed to classify EEG MI signals. This architecture comprised three primary blocks: the GCN block, the Global Mean Pooling Block, and the Fully Connected Block. In the GCN Block, generalised graph features for each EEG electrode were extracted. Subsequently, the features from all 64 channels were consolidated using a mean in the Global Mean Pooling Block. The Fully Connected Block was employed for the final prediction. A detailed representation of this model architecture is provided in Figure~\ref{fig: overall_archi} and Table~\ref{table: model archi}.

\begin{table*}[!ht]
\caption{Generalised Architecture of GCN Model}
\label{table: model archi}
\centering
\setlength{\tabcolsep}{3pt}
\begin{tabular}{m{1.5cm}<{\centering} m{3.0cm} <{\centering} m{1.5cm}<{\centering} m{1.5cm}<{\centering} m{2.5cm}<{\centering} m{1.5cm}<{\centering} m{1.5cm}<{\centering} m{1.5cm}<{\centering} }
\hline
\hline
Layer & Type & Input Size & Polynomial Order & Weights & Bias & Output & Activation \\ \hline
Input & Input & $N\times 1$ & - & - & - & - & - \\ 
\hline
\multicolumn{8}{c}{Block A - GCN Block} \\
\hline
C1 & Graph Convolution & $N\times 1$ & $K_1$ & $1\times F_1 \times K_1$ & $N\times F_1$ & $N\times F_1$ & - \\
BNC1 & Batch Normalisation & $N\times F_1$ & - & $F_1$ & $F_1$ & $N\times F_1$ & ReLU \\
C2 & Graph Convolution & $N\times F_1$ & $K_2$ & $F_1\times F_2 \times K_2$ & $N\times F_2$ & $N\times F_2$ & - \\
BNC2 & Batch Normalisation & $N\times F_2$ & - & $F_2$ & $F_2$ & $N\times F_2$ & ReLU \\
C3 & Graph Convolution & $N\times F_2$ & $K_3$ & $F_2\times F_3 \times K_3$ & $N\times F_3$ & $N\times F_3$ & - \\
BNC3 & Batch Normalisation & $N\times F_3$ & - & $F_3$ & $F_3$ & $N\times F_3$ & ReLU \\
C4 & Graph Convolution & $N\times F_3$ & $K_4$ & $F_3\times F_4 \times K_4$ & $N\times F_4$ & $N\times F_4$ & - \\
BNC4 & Batch Normalisation & $N\times F_4$ & - & $F_4$ & $F_4$ & $N\times F_4$ & ReLU \\
C5 & Graph Convolution & $N\times F_4$ & $K_5$ & $F_4\times F_5 \times K_5$ & $N\times F_5$ & $N\times F_5$ & - \\
BNC5 & Batch Normalisation & $N\times F_5$ & - & $F_5$ & $F_5$ & $N\times F_5$ & ReLU \\
C6 & Graph Convolution & $N\times F_5$ & $K_6$ & $F_5\times F_6 \times K_6$ & $N\times F_6$ & $N\times F_6$ & - \\
BNC6 & Batch Normalisation & $N\times F_6$ & - & $F_6$ & $F_6$ & $N\times F_6$ & ReLU \\
\hline
\multicolumn{8}{c}{Block B - Global Mean Pooling Block} \\
\hline
P & Global Mean Pool & $N\times F_6$ & - & - & - & $F_6$ & - \\
\hline
\multicolumn{8}{c}{Block C - Fully Connected Block} \\
\hline
FC1 & Fully Connected & $F_6$ & - & $F_6 \times H_1$ & $H_1$ & $H_1$ & - \\
BNFC1 & Batch Normalisation & $H_1$ & - & $H_1$ & $H_1$ & $H_1$ & ReLU \\
FC2 & Fully Connected & $H_1$ & - & $H_1 \times H_2$ & $H_2$ & $H_2$ & - \\
BNFC2 & Batch Normalisation & $H_2$ & - & $H_2$ & $H_2$ & $H_2$ & ReLU \\
FC3 & Fully Connected & $H_2 \times O$ & - & $H_2 \times O$ & $O$ & $O$ & - \\
S & Softmax Classification & $O$ & - & - & - & $O$ & - \\
\hline
\hline
\multicolumn{8}{c}{$N= $ Number of EEG Channels (i.e. 64); $O= $ Number of EEG MI Classes (i.e. 4)} \\

\end{tabular}
\end{table*}

\subsection{Model Setting}

Let $F_i$ represent the number of filters at each GCN level, given by  $F_i \in [F_1, F_2, F_3, F_4, F_5, F_6]$. Similarly, $K_i$ denotes the polynomial order of the filter for each $i^{th}$ layer, and is defined as $K_i \in [K_1, K_2, K_3, K_4, K_5, K_6]$. $O$ indicates the number of MI classes for prediction. Due to the large volume of instances in the training set, we employed a mini-batch size $B$ of 1024. A batch normalisation (BN) layer was incorporated after both the spectral GCN and Fully Connected layers. This BN layer re-scales and re-centers normalised signals to match the original distribution within the mini-batch, addressing the internal covariate shift issue and helping to mitigate the gradient vanishing/exploding problem. Additionally, 50\% dropout layers were integrated after the ReLU layers (Equation~\ref{eqn: relu}) within the Fully Connected Block for regularisation. The details of the model settings can be found in Table~\ref{table: model set}, while the hyperparameter settings are provided in Table~\ref{table: hyperparamter}.

\begin{equation}
\label{eqn: relu}
ReLU(x)=max(0, x)
\end{equation}

\begin{equation}
\label{eqn: softmax}
Softmax(\hat{y_i}) = \frac{e^{\hat{y_i}}}{\sum_{i=1}^{O}e^{\hat{y_i}}}
\end{equation}

\noindent where $\hat{y_i}$ represent the predicted probability of an instance for each class, ranging over $\hat{y_i}\in[\hat{y_1}, ..., \hat{y_O}]$. $O$ denotes the total number of classes. The loss function employed was the cross-entropy loss.

\begin{equation}
\label{eqn: cross-entropy}
Loss=-\frac{1}{|B|}\sum_{b=1}^{B}\sum_{i=1}^{O}y_i . log(\hat{y_i})
\end{equation}

\begin{table}[ht]
\caption{Model Setting}
\label{table: model set}
\centering
\setlength{\tabcolsep}{3pt}
\begin{tabular}{m{0.5cm}<{\centering} m{2.7cm} <{\centering} m{1.5cm}<{\centering} m{1.5cm}<{\centering} m{1.5cm}<{\centering}}
\hline
\hline
Model & Model Framework & Number of GCN Filters & GCN Filter Polynomial Order & Number of FC Hidden Nodes\\
\hline
A & $(C-BNC)\times6-P-(FC-BNFC)\times2-FC-S$ & 16, 32, 64, 128, 256, 512 & 5, 5, 5, 5, 5, 5 & 1024, 2048, 4 \\
\hline
B & $(C-BNC)\times6-P-(FC-BNFC)\times2-FC-S$ & 16, 32, 64, 128, 256, 512 & 2, 2, 2, 2, 2, 2 & 1024, 2048, 4 \\
\hline
C & $(C-BNC)\times5-P-(FC-S)$ & 16, 32, 64, 128, 256 & 5, 5, 5, 5, 5 & 4 \\
\hline
D & $(C-BNC)\times5-P-(FC-S)$ & 16, 32, 64, 128, 256 & 2, 2, 2, 2, 2 & 4 \\
\hline
E & $(C-BNC)\times5-P-(FC-BNFC)\times2-FC-S$ & 64, 128, 256, 512, 1024 & 5, 5, 5, 5, 5 & 512, 128, 4 \\
\hline
F & $(C-BNC)\times5-P-(FC-BNFC)\times2-FC-S$ & 64, 128, 256, 512, 1024 & 2, 2, 2, 2, 2 & 512, 128, 4\\
\hline

\end{tabular}
\end{table}

\begin{table}[!ht]
\caption{Hyperparameter Setting}
\label{table: hyperparamter}
\centering
\setlength{\tabcolsep}{3pt}
\begin{tabular}{m{3.5cm}<{\centering} m{1.5cm} <{\centering}}
\hline
\hline
Hyperparamter & Value \\
\hline
Training Epochs $(N_{ep})$ & 1000 \\
Batch Size $(B)$ & 1024 \\
Dropout Rate & 0.5 \\
Optimizer & Adam \\
Initial Learning Rate $(\eta)$ & 0.01 \\
\hline
\hline

\end{tabular}
\end{table}

Both accuracy and F1 score evaluation metrics were employed to assess the performance of models.
\begin{equation}
\label{eqn: accuracy}
Accuracy=\frac{TP + TN}{TP + FP + TN + FN}
\end{equation}

\begin{equation}
\label{eqn: sensitivity}
Sensitivity=\frac{TP}{TP + FN}
\end{equation}

\begin{equation}
\label{eqn: precision}
Precision=\frac{TP}{TP + FP}
\end{equation}

\begin{equation}
\label{eqn: f1 score}
F1\ Score=\frac{2\times Precision\times Sensitivity}{Precision + Sensitivity}
\end{equation}

\begin{figure}
    \centering
    \includegraphics[width=0.20\textwidth]{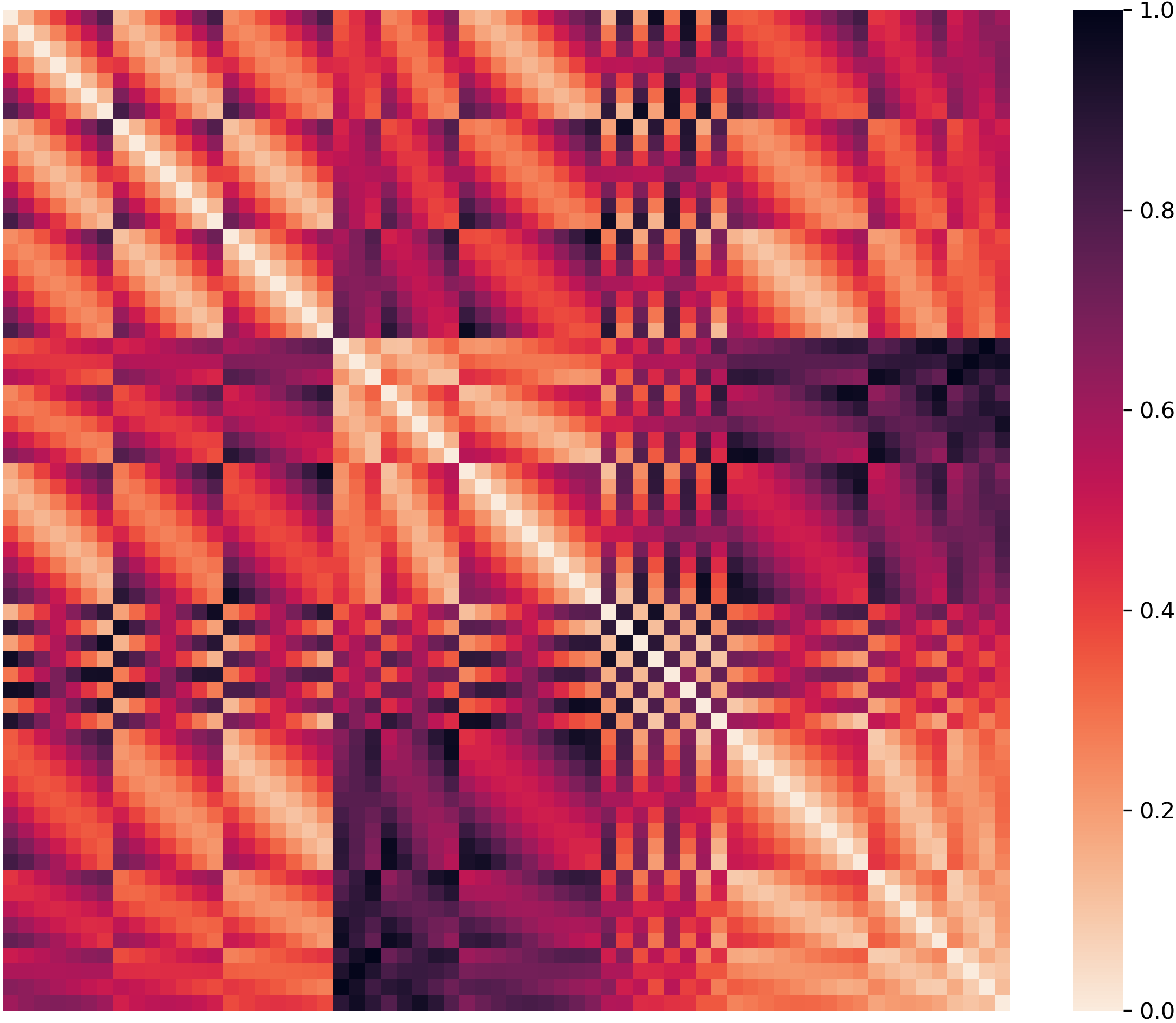}
    \caption{Geodesic Distance Adjacency Matrix ($A^{Geodesic}$)}
    \label{fig: adj_geodesic_subj6_subj14}

\end{figure}

\begin{figure}
    \centering
    \begin{subfigure}[t]{0.20\textwidth}
        \centering
        \includegraphics[width=1\textwidth]{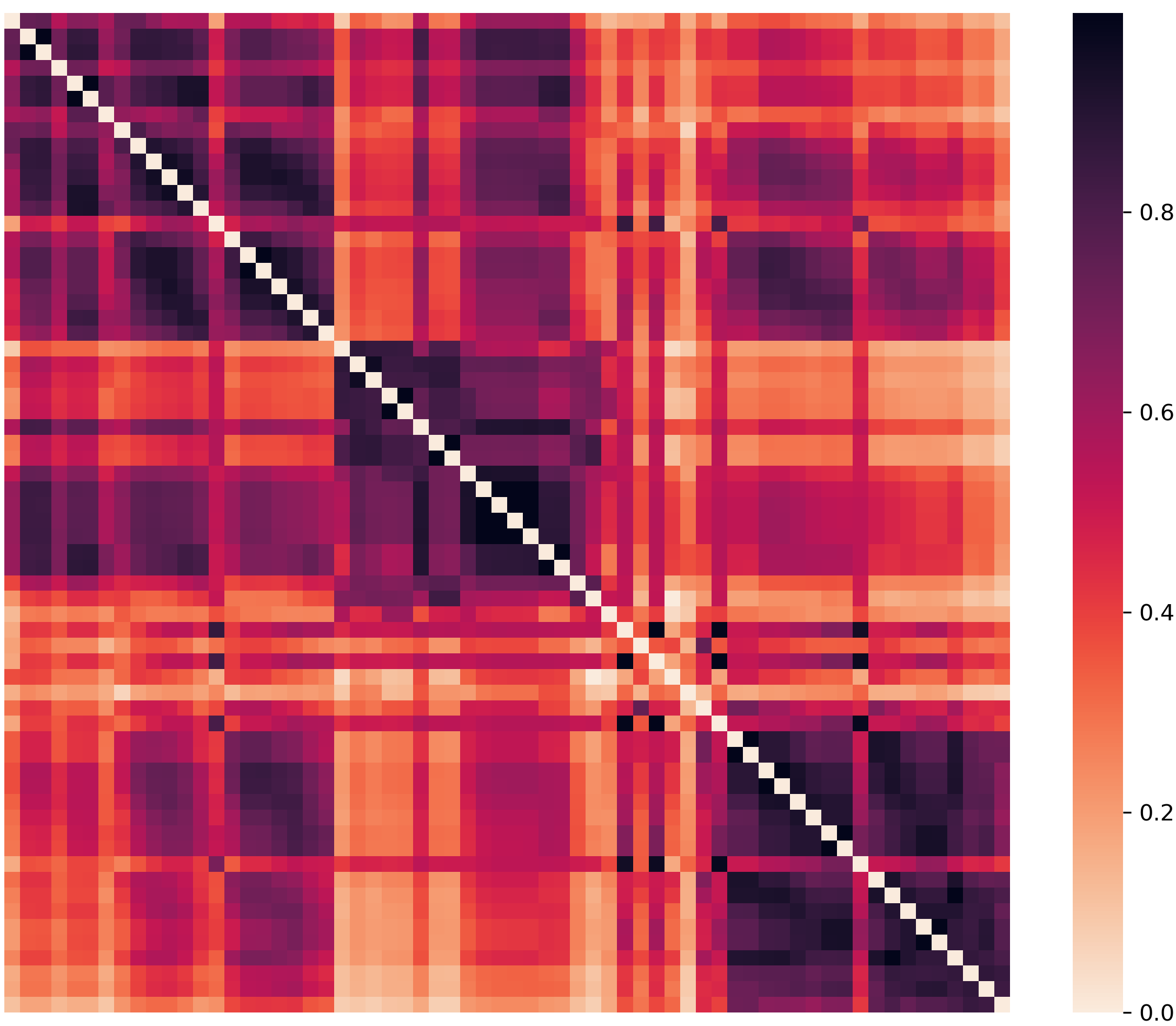}
        \caption{Subject $S_6$}
        \label{fig: adj_pcc_subj6}
    \end{subfigure}
    \hfill
    \begin{subfigure}[t]{0.20\textwidth}
        \centering
        \includegraphics[width=1\textwidth]{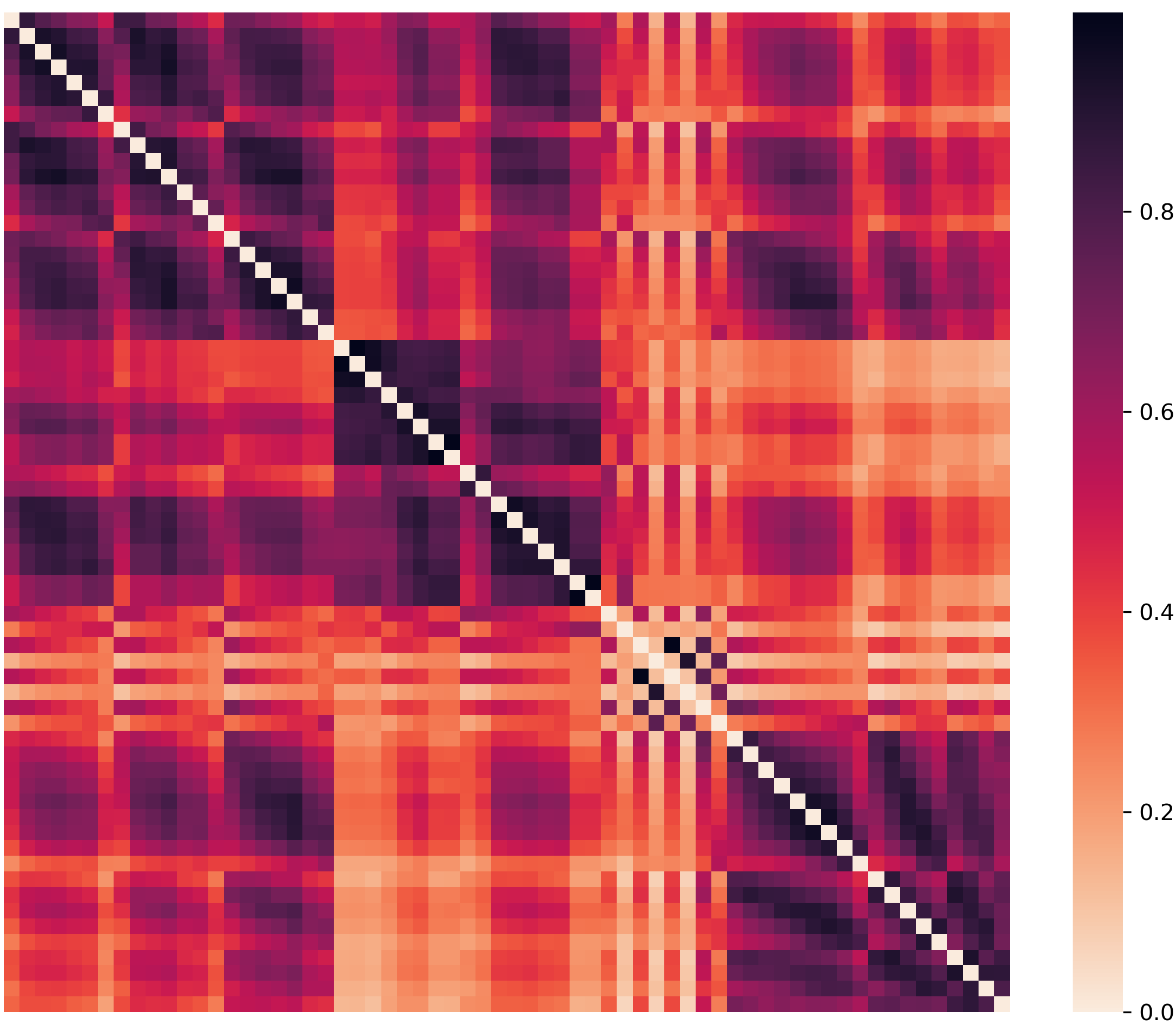}
        \caption{Subject $S_{14}$}
        \label{fig: adj_pcc_subj14}        
    \end{subfigure}
    \caption{PCC Adjacency Matrix ($A^{PCC}$) of Subject $S_6$ and $S_{14}$}
    \label{fig: adj_pcc_subj6_subj14}
\end{figure}

\begin{figure}
    \centering
    \begin{subfigure}[t]{0.22\textwidth}
        \centering
        \includegraphics[width=1\textwidth]{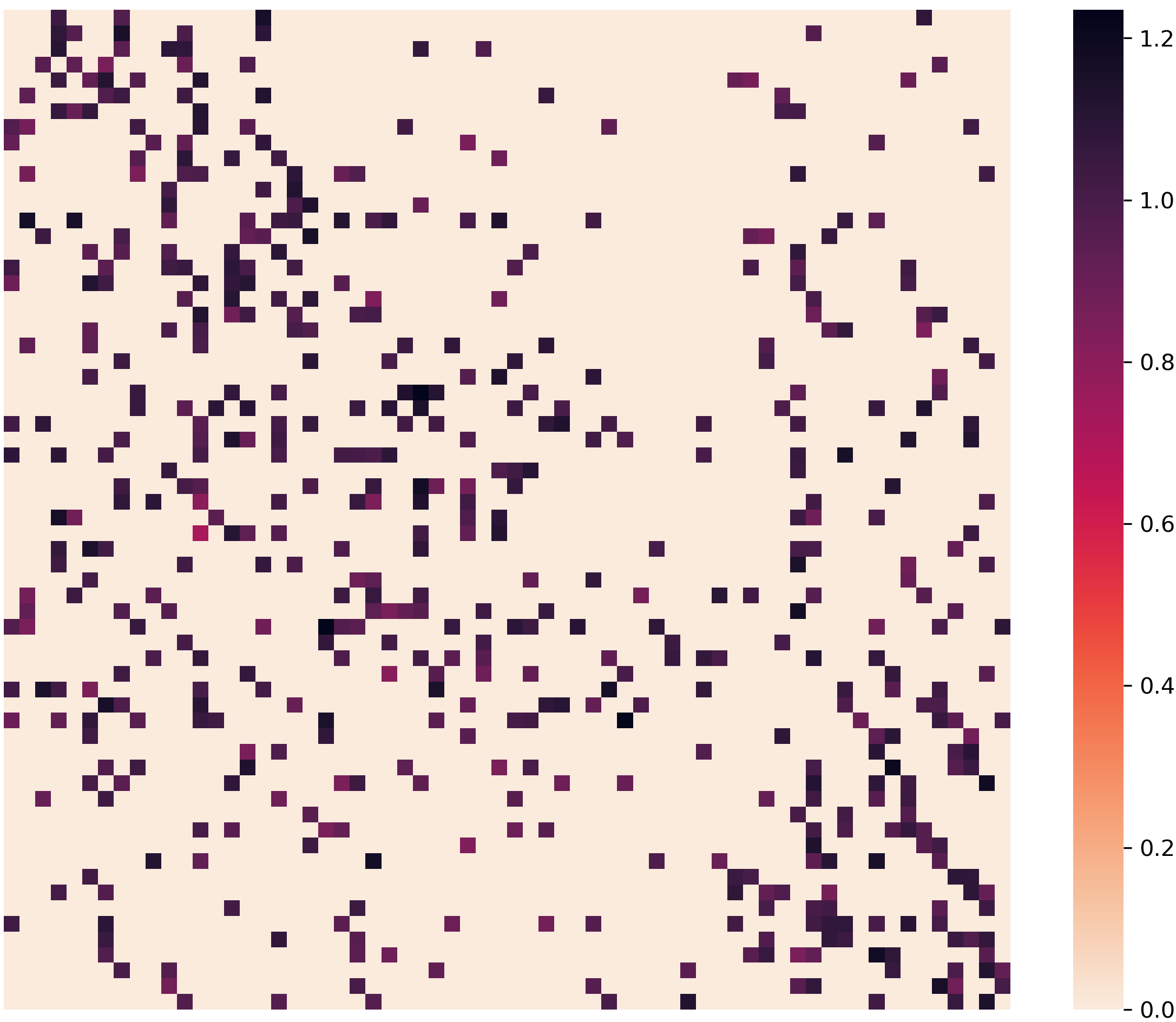}
        \caption{Adjacency Matrix ($m_{g\_EEG\_GLT}$) - Subject $S_6$, 13.39\% Density, Model A}
        \label{fig: adj_glt_subj6_modelA}
    \end{subfigure}
    \hfill
    \begin{subfigure}[t]{0.22\textwidth}
        \centering
        \includegraphics[width=1\textwidth]{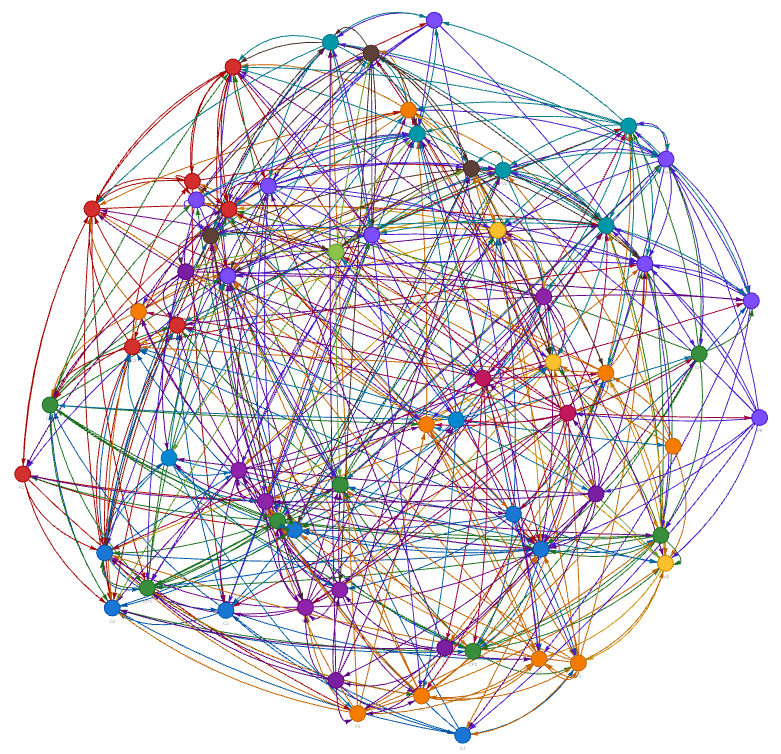}
        \caption{Graph of $m_{g\_EEG\_GLT}$ - $S_6$, 13.39\% Density, Model A}
        \label{fig: node_edges_subj6_modelA_19}        
    \end{subfigure}

    \vspace{1em} 

    \centering
    \begin{subfigure}[t]{0.22\textwidth}
        \centering
        \includegraphics[width=1\textwidth]{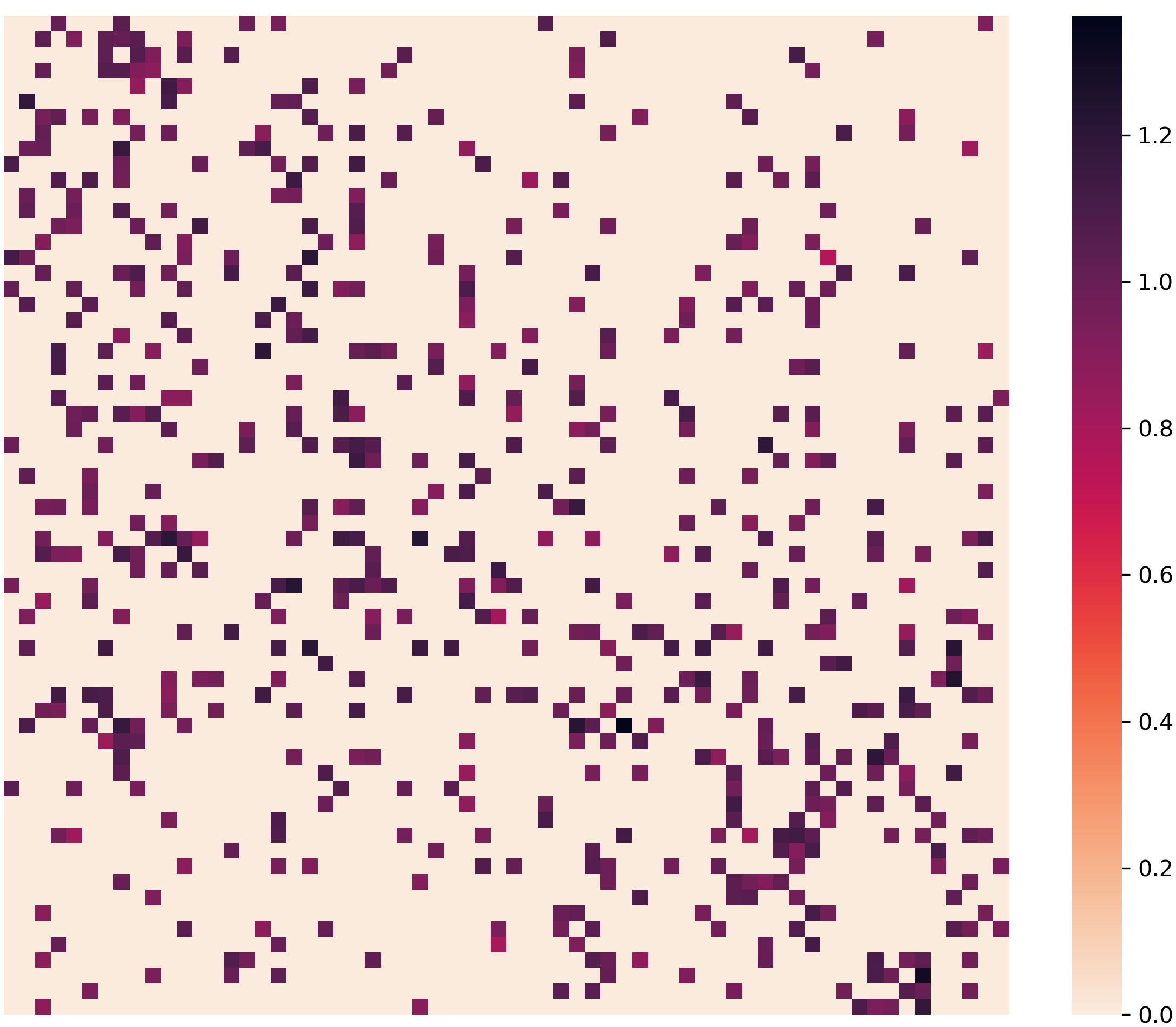}
        \caption{Adjacency Matrix ($m_{g\_EEG\_GLT}$) - Subject $S_6$, 13.39\% Density, Model E}
        \label{fig: adj_glt_subj6_modelE}
    \end{subfigure}
    \hfill
    \begin{subfigure}[t]{0.22\textwidth}
        \centering
        \includegraphics[width=1\textwidth]{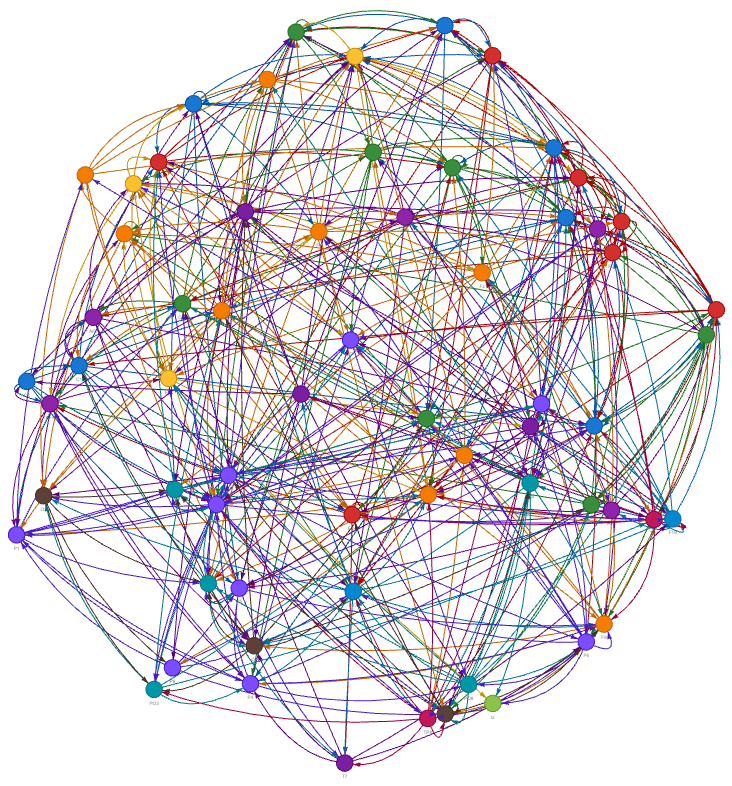}
        \caption{Graph of $m_{g\_EEG\_GLT}$ - $S_6$, 13.39\% Density, Model E}
        \label{fig: node_edges_subj6_modelE_19}        
    \end{subfigure}
    \caption{Representations of $m_{g\_EEG\_GLT}$ for Subject $S_6$ at 13.39\% Density. 
    (a) Adjacency Matrix - Model A (Accuracy: 78.13\%)
    (b) Graph - Model A
    (c) Adjacency Matrix - Model E (Accuracy: 73.55\%)
    (d) Graph - Model E}
    \label{fig: adj_glt_subj6_subj14_modelA_modelE}
\end{figure}

\section{Results and Discussion}

\begin{table*}[ht]
\caption{Accuracy Comparison Across Different Methods of Adjacency Matrix Construction for Each Subject}
\label{table: Subj_accuracy_comparison}
\centering
\setlength{\tabcolsep}{3pt}
\begin{tabular}{m{1.0cm}<{\centering} m{2.2cm}<{\centering} m{2.2cm}<{\centering} m{2.2cm}<{\centering}
m{0.1cm}<{\centering}    m{2.2cm}<{\centering} m{2.2cm}<{\centering} m{2.2cm}<{\centering}}
\hline
\hline
\multirow{2}{*}{Subject} & \multicolumn{3}{c}{Accuracy (Mean$\pm$Std)} & & \multicolumn{3}{c}{F1 Score (Mean$\pm$Std)}\\
\cline{2-4} \cline{6-8}
 & Geodesic & PCC & EEG\_GLT \newline(our method) & & Geodesic & PCC & EEG\_GLT \newline(our method) \\
\hline

$S_{1}$ & 66.19\% $\pm$ 4.17\% & 76.47\% $\pm$ 9.94\% & \textbf{98.51\% $\pm$ 0.77\%}
        & & 66.53\% $\pm$ 4.36\% & 76.91\% $\pm$ 9.78\% & \textbf{98.53\% $\pm$ 0.78\%} \\
$S_{2}$ & 46.53\% $\pm$ 1.33\% & 69.13\% $\pm$ 7.05\% & \textbf{76.18\% $\pm$ 5.53\%} 
        & & 46.47\% $\pm$ 1.46\% & 69.34\% $\pm$ 7.37\% & \textbf{76.19\% $\pm$ 5.52\%} \\
$S_{3}$ & 76.18\% $\pm$ 4.98\% & 87.28\% $\pm$ 9.19\% & \textbf{99.17\% $\pm$ 0.32\%} 
        & & 76.12\% $\pm$ 5.00\% & 87.43\% $\pm$ 8.97\% & \textbf{99.19\% $\pm$ 0.31\%} \\
$S_{4}$ & 96.41\% $\pm$ 1.97\% & 99.13\% $\pm$ 1.01\% & \textbf{99.97\% $\pm$ 0.06\%} 
        & & 96.44\% $\pm$ 1.98\% & 99.10\% $\pm$ 1.12\% & \textbf{99.97\% $\pm$ 0.05\%} \\
$S_{5}$ & 37.05\% $\pm$ 1.04\% & 43.19\% $\pm$ 3.03\% & \textbf{50.95\% $\pm$ 3.80\%} 
        & & 36.66\% $\pm$ 0.97\% & 43.28\% $\pm$ 2.73\% & \textbf{50.86\% $\pm$ 3.85\%} \\
$S_{6}$ & 44.37\% $\pm$ 1.59\% & 58.23\% $\pm$ 5.19\% & \textbf{69.60\% $\pm$ 5.67\%} 
        & & 44.29\% $\pm$ 1.65\% & 58.25\% $\pm$ 5.49\% & \textbf{69.50\% $\pm$ 5.70\%} \\
$S_{7}$ & 40.44\% $\pm$ 1.19\% & 50.98\% $\pm$ 3.80\% & \textbf{59.45\% $\pm$ 3.00\%} 
        & & 40.30\% $\pm$ 1.23\% & 51.10\% $\pm$ 3.49\% & \textbf{59.34\% $\pm$ 2.99\%} \\
$S_{8}$ & 89.03\% $\pm$ 7.04\% & 95.06\% $\pm$ 5.96\% & \textbf{99.95\% $\pm$ 0.07\%} 
        & & 88.84\% $\pm$ 6.88\% & 95.14\% $\pm$ 5.81\% & \textbf{99.96\% $\pm$ 0.07\%} \\
$S_{9}$ & 87.26\% $\pm$ 14.26\% & 97.64\% $\pm$ 3.33\% & \textbf{99.95\% $\pm$ 0.08\%} 
        & & 87.41\% $\pm$ 14.49\% & 97.70\% $\pm$ 3.78\% & \textbf{99.95\% $\pm$ 0.08\%} \\
$S_{10}$ & 98.26\% $\pm$ 0.31\% & 99.24\% $\pm$ 0.19\% & \textbf{99.99\% $\pm$ 0.01\%} 
        & & 98.25\% $\pm$ 0.32\% & 99.25\% $\pm$ 0.20\% & \textbf{99.99\% $\pm$ 0.01\%} \\
$S_{11}$ & 97.18\% $\pm$ 1.12\% & 99.48\% $\pm$ 0.70\% & \textbf{99.99\% $\pm$ 0.01\%} 
        & &97.18\% $\pm$ 1.13\% & 99.49\% $\pm$ 0.74\% & \textbf{99.99\% $\pm$ 0.01\%} \\
$S_{12}$ & 71.54\% $\pm$ 3.44\% & 78.07\% $\pm$ 8.95\% & \textbf{99.69\% $\pm$ 0.32\%} 
        & & 71.40\% $\pm$ 3.37\% & 77.94\% $\pm$ 8.76\% & \textbf{99.70\% $\pm$ 0.31\%} \\
$S_{13}$ & 36.52\% $\pm$ 0.32\% & 41.35\% $\pm$ 1.23\% & \textbf{44.50\% $\pm$ 2.23\%} 
        & & 36.49\% $\pm$ 0.45\% & 41.01\% $\pm$ 1.34\% & \textbf{44.47\% $\pm$ 2.23\%} \\
$S_{14}$ & 40.21\% $\pm$ 1.80\% & 55.97\% $\pm$ 6.47\% & \textbf{72.39\% $\pm$ 6.43\%} 
        & & 40.10\% $\pm$ 1.88\% & 56.05\% $\pm$ 6.57\% & \textbf{72.71\% $\pm$ 6.13\%} \\
$S_{15}$ & 46.16\% $\pm$ 1.28\% & 52.11\% $\pm$ 3.96\% & \textbf{67.55\% $\pm$ 9.26\%} 
        & & 45.92\% $\pm$ 1.93\% & 52.20\% $\pm$ 3.66\% & \textbf{67.52\% $\pm$ 9.27\%} \\
$S_{16}$ & 95.62\% $\pm$ 3.87\% & 96.75\% $\pm$ 5.00\% & \textbf{99.98\% $\pm$ 0.03\%} 
        & & 94.94\% $\pm$ 5.25\% & 96.72\% $\pm$ 5.07\% & \textbf{99.98\% $\pm$ 0.03\%} \\
$S_{17}$ & 92.07\% $\pm$ 8.10\% & 98.83\% $\pm$ 2.33\% & \textbf{99.98\% $\pm$ 0.03\% }
        & & 91.95\% $\pm$ 8.31\% & 98.66\% $\pm$ 2.76\% & \textbf{99.98\% $\pm$ 0.03\%} \\
$S_{18}$ & 71.24\% $\pm$ 5.96\% & 86.19\% $\pm$ 10.95\% & \textbf{99.92\% $\pm$ 0.12\%} 
        & & 73.28\% $\pm$ 3.28\% & 85.98\% $\pm$ 11.10\% & \textbf{99.93\% $\pm$ 0.13\%} \\
$S_{19}$ & 33.18\% $\pm$ 0.40\% & 38.38\% $\pm$ 2.27\% & \textbf{41.41\% $\pm$ 1.44\%} 
        & & 32.85\% $\pm$ 0.32\% & 38.35\% $\pm$ 2.32\% & \textbf{41.27\% $\pm$ 1.34\%} \\
$S_{20}$ & 93.77\% $\pm$ 2.08\% & 98.44\% $\pm$ 0.68\% & \textbf{99.94\% $\pm$ 0.11\% }
        & & 93.76\% $\pm$ 2.06\% & 98.45\% $\pm$ 0.72\% & \textbf{99.95\% $\pm$ 0.12\%} \\

\hline
\hline
\end{tabular}
\end{table*}

\subsection{Geodesic vs PCC Adjacency Matrix Construction Method}

The Table IV presents the mean performance accuracy and F1 score across various models for different adjacency matrix construction methods, including Geodesic, PCC, and EEG\_GLT, for each subject. Among the existing methods (PCC and Geodesic), the PCC adjacency method consistently outperformed the Geodesic method, enhancing the accuracy by 0.98\% - 22.60\% and the F1 score by 0.99\% - 22.86\%. Table V and Figure 6 detail the mean accuracies and F1 scores for 20 subjects ($S_1$ - $S_{20}$) across different matrix construction methods for each model setting. Notably, the PCC method outperformed the Geodesic method across all model settings, improving accuracy by 9.76\% and the F1 score by 9.63\%. The superiority of the PCC method in EEG MI adjacency matrix construction over the Geodesic method stems a major limitation in the latter: it considers only the geodesic distance between EEG electrodes, leading to identical adjacency matrices for all 20 subjects (Figure~\ref{fig: adj_geodesic_subj6_subj14}). In contrast, the PCC method produces unique matrices for each subject, offering tailored matrices that are better suited for subject-based EEG MI classification (Figure~\ref{fig: adj_pcc_subj6_subj14}).

Our experiment revealed that using the relative physical distance between EEG electrodes was suboptimal due to limited accuracy. Since EEG electrodes do not have direct connections to brain tissue, electrical signals produced by large neuron groups that fire simultaneously or synchronously need to traverse multiple tissue layers such as the cerebral cortex, cerebrospinal fluid, skull, and scalp before detected by EEG electrodes. Given that the skull attenuates these signals, and causes a smearing effect \cite{van1998volume}, coupled with individual differences in skull thickness, scalp conductivity, and MI task approach, it was the most logical to use unique adjacency matrices for each individual.

In the $A^{Geodesic}$ adjacency matrix construction, we adopted a unit sphere assumption because the PhysioNet dataset lacks data on individual head shapes. Given natural variations in head structure, $A^{Geodesic}$ values could potentially differ for each subject.

\begin{table}[ht]
\caption{Accuracy Comparison Across Different Methods of Adjacency Matrix Construction for Each Model}
\label{table: Model_accuracy_comparison}
\centering
\setlength{\tabcolsep}{2pt}
\begin{tabular}{m{1.5cm}<{\centering} m{1.5cm}<{\centering} m{1.8cm}<{\centering} m{1.8cm}<{\centering}}
\hline
\hline
Model & Adj Method & Avg. Accuracy & Avg. F1 Score \\
\hline
\multirow{3}{*}{Model A} & Geodesic  & 70.70\% & 70.14\% \\
                         & PCC       & 79.82\% & 79.77\% \\
                         & EEG\_GLT  & \textbf{85.90\%} & \textbf{85.89\%} \\
\hline
\multirow{3}{*}{Model B} & Geodesic  & 70.70\% & 70.65\% \\
                         & PCC       & 78.69\% & 78.32\% \\
                         & EEG\_GLT  & \textbf{83.84\%} & \textbf{83.80\%} \\
\hline
\multirow{3}{*}{Model C} & Geodesic  & 65.49\% & 65.43\% \\
                         & PCC       & 74.13\% & 74.41\% \\
                         & EEG\_GLT  & \textbf{83.27\%} & \textbf{83.28\%} \\
\hline
\multirow{3}{*}{Model D} & Geodesic  & 62.97\% & 63.08\% \\
                         & PCC       & 68.13\% & 68.05\% \\
                         & EEG\_GLT  & \textbf{81.52\%} & \textbf{81.48\%} \\
\hline
\multirow{3}{*}{Model E} & Geodesic  & 69.20\% & 69.16\% \\
                         & PCC       & 78.90\% & 78.88\% \\
                         & EEG\_GLT  & \textbf{85.91\%} & \textbf{85.88\%} \\
\hline
\multirow{3}{*}{Model F} & Geodesic  & 69.34\% & 69.28\% \\
                         & PCC       & 76.89\% & 77.26\% \\
                         & EEG\_GLT  & \textbf{83.26\%} & \textbf{83.36\%} \\
\hline
\hline
\end{tabular}
\end{table}

\begin{figure}
    \centering
    \includegraphics[width=0.48\textwidth]{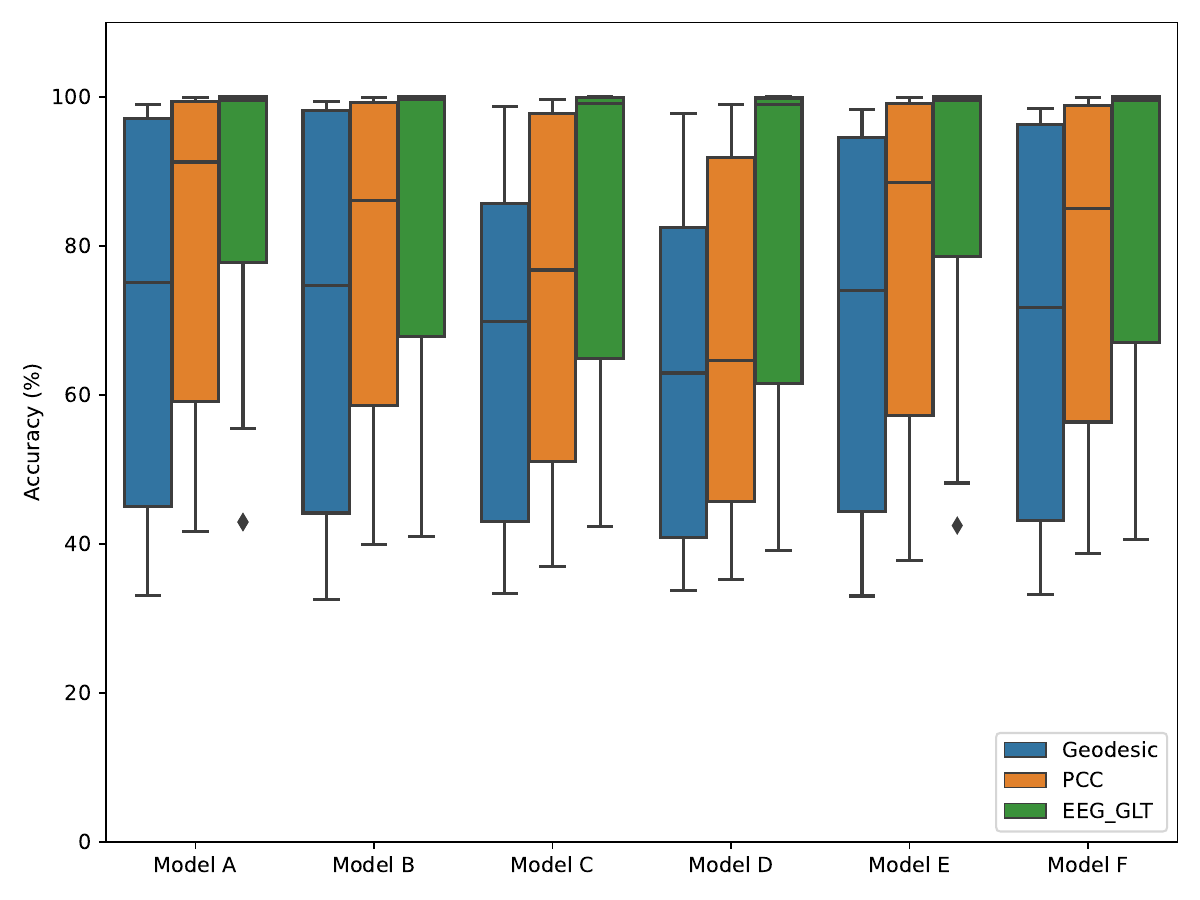}
    \caption{Comparison of Model Accuracy Across Different Adjacency Matrix Construction Methods}
    \label{fig: box_plot_models}

\end{figure}

\subsection{EEG\_GLT Method vs PCC Method in Adjacency Matrix Construction}
\label{subsec: eeg_vs_pcc}

Our EEG\_GLT method consistently surpassed the PCC method in both accuracy and F1 score. As shown in Table~\ref{table: Subj_accuracy_comparison}, EEG\_GLT demonstrated substantial increase in accuracy and F1 score compared to the PCC method, by 0.52\% - 22.04\% and 0.50\% - 21.76\%, respectively. Unlike the PCC method, our EEG\_GLT adjacency matrix is dynamic with the ability to adapt to both the individual subject and the model settings of GCNs (Table~\ref{table: model set}), as shown in Figure~\ref{fig: adj_glt_subj6_subj14_modelA_modelE}. 

According to Table~\ref{table: Model_accuracy_comparison} and Figure~\ref{fig: box_plot_models}, our EEG\_GLT method improved the mean accuracies and F1 scores for 20 subjects by 13.39\% and 13.43\%, respectively compared to the PCC method. This underscores the necessity of model-specific adjustments, in addition to subject-based tailoring in the adjacency matrix construction, to attain the best possible outcomes. Distinctly, our EEG\_GLT matrix is asymmetrical due to the iterative pruning process detailed in Algorithm~\ref{alg: EEG_GLT}, which refines the matrix until the optimal EEG Graph Lottery Ticket is identified. 

\begin{figure*}
    \centering
    \begin{subfigure}[t]{0.32\textwidth}
        \centering
        \includegraphics[width=1\textwidth]{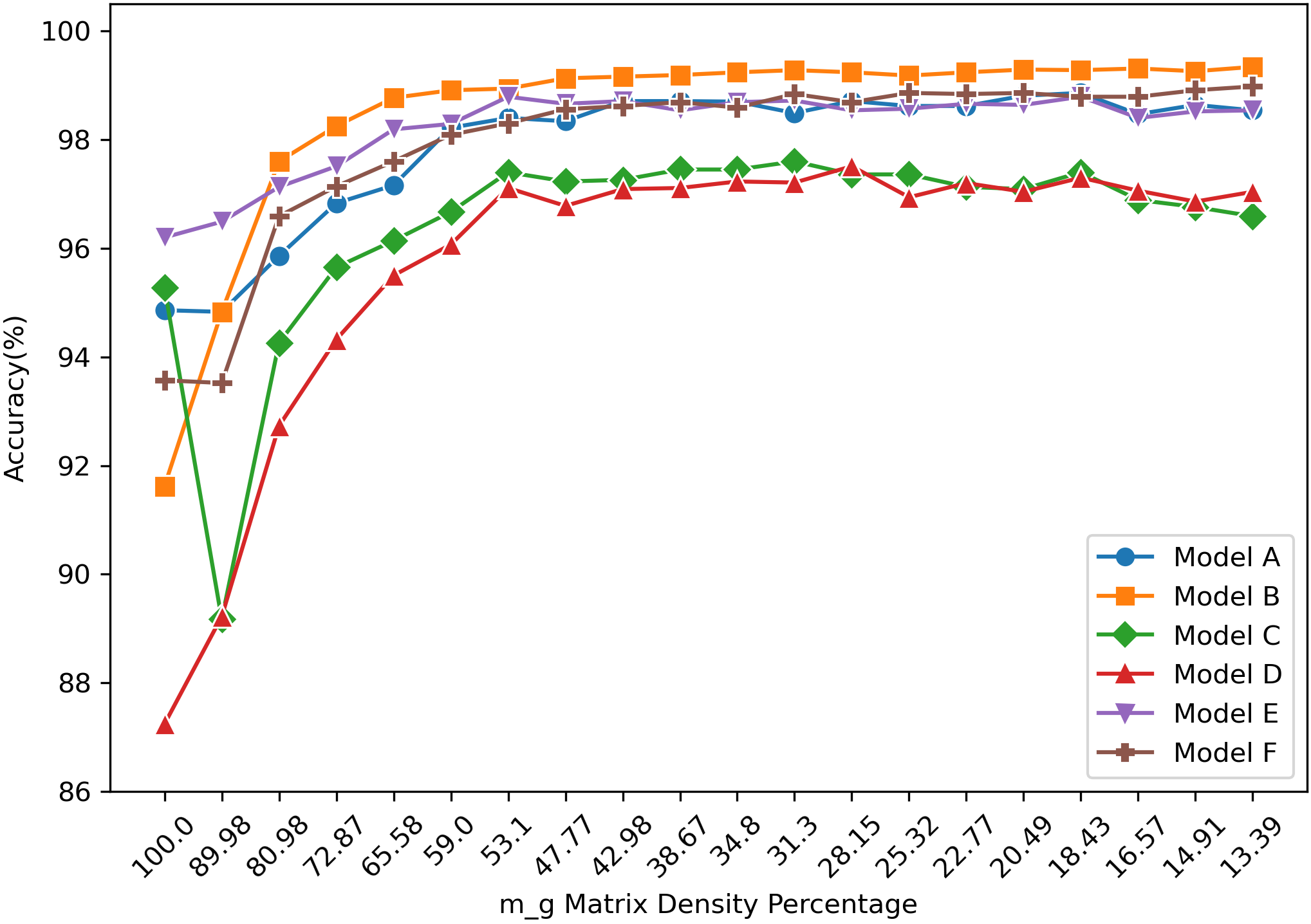}
        \caption{Subject $S_1$}
        \label{fig: Subj-1-acc}
    \end{subfigure}
    \hfill
    \begin{subfigure}[t]{0.32\textwidth}
        \centering
        \includegraphics[width=1\textwidth]{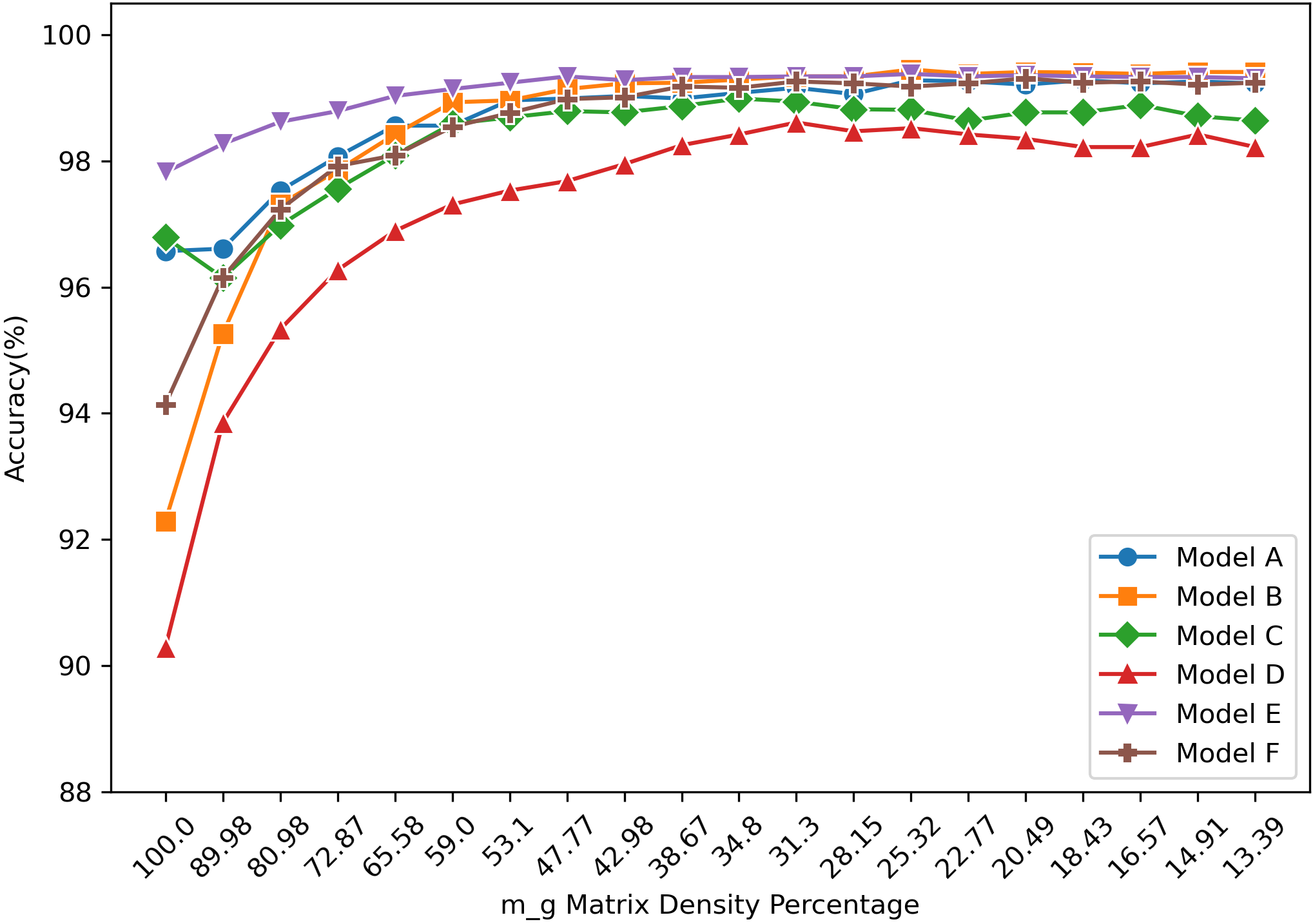}
        \caption{Subject $S_3$}
        \label{fig: Subj-3-acc}        
    \end{subfigure}
    \hfill
    \begin{subfigure}[t]{0.32\textwidth}
        \centering
        \includegraphics[width=1\textwidth]{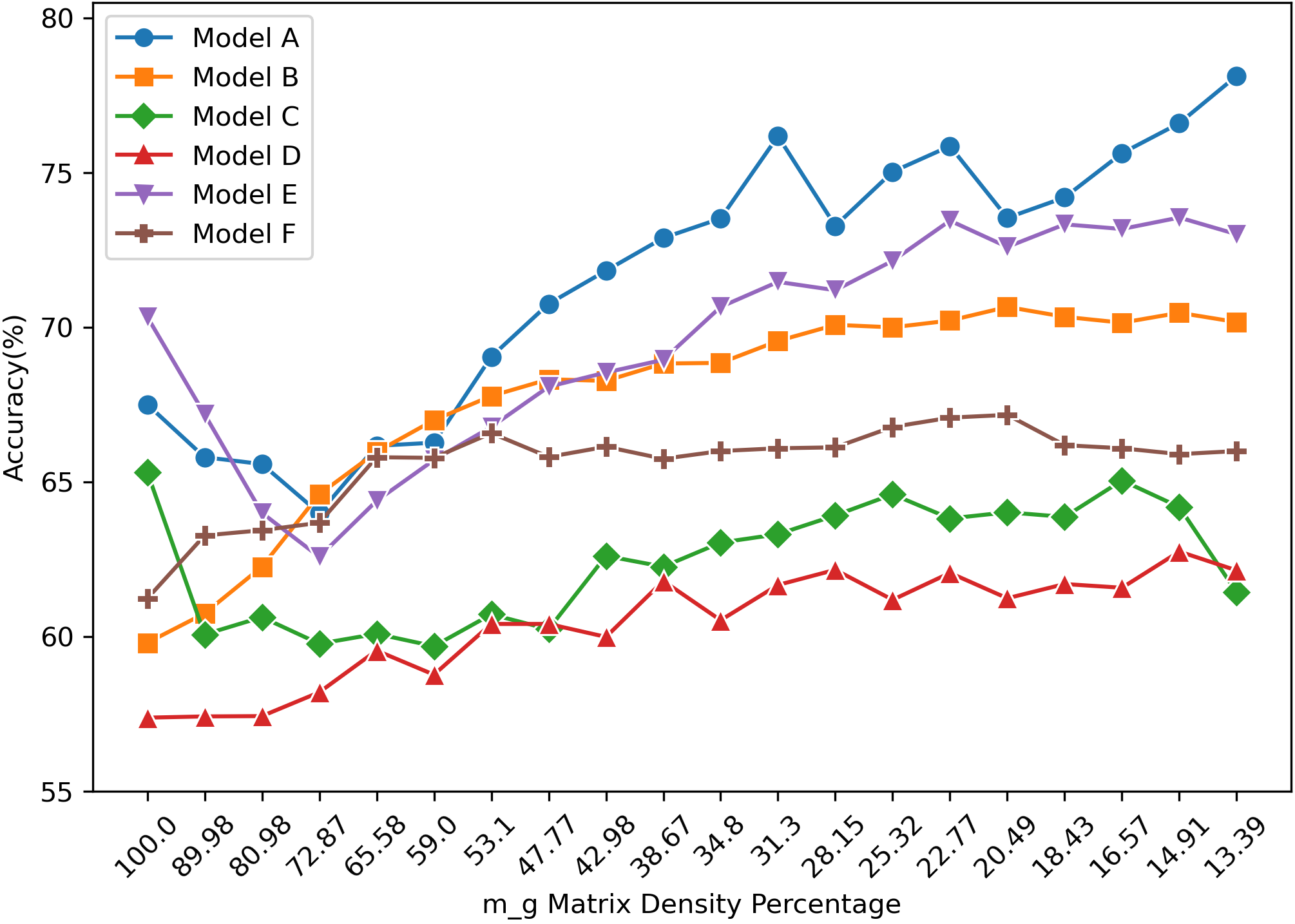}
        \caption{Subject $S_6$}
        \label{fig: Subj-6-acc}
    \end{subfigure}
    
    \vspace{1em} 
    
    \begin{subfigure}[t]{0.32\textwidth}
        \centering
        \includegraphics[width=1\textwidth]{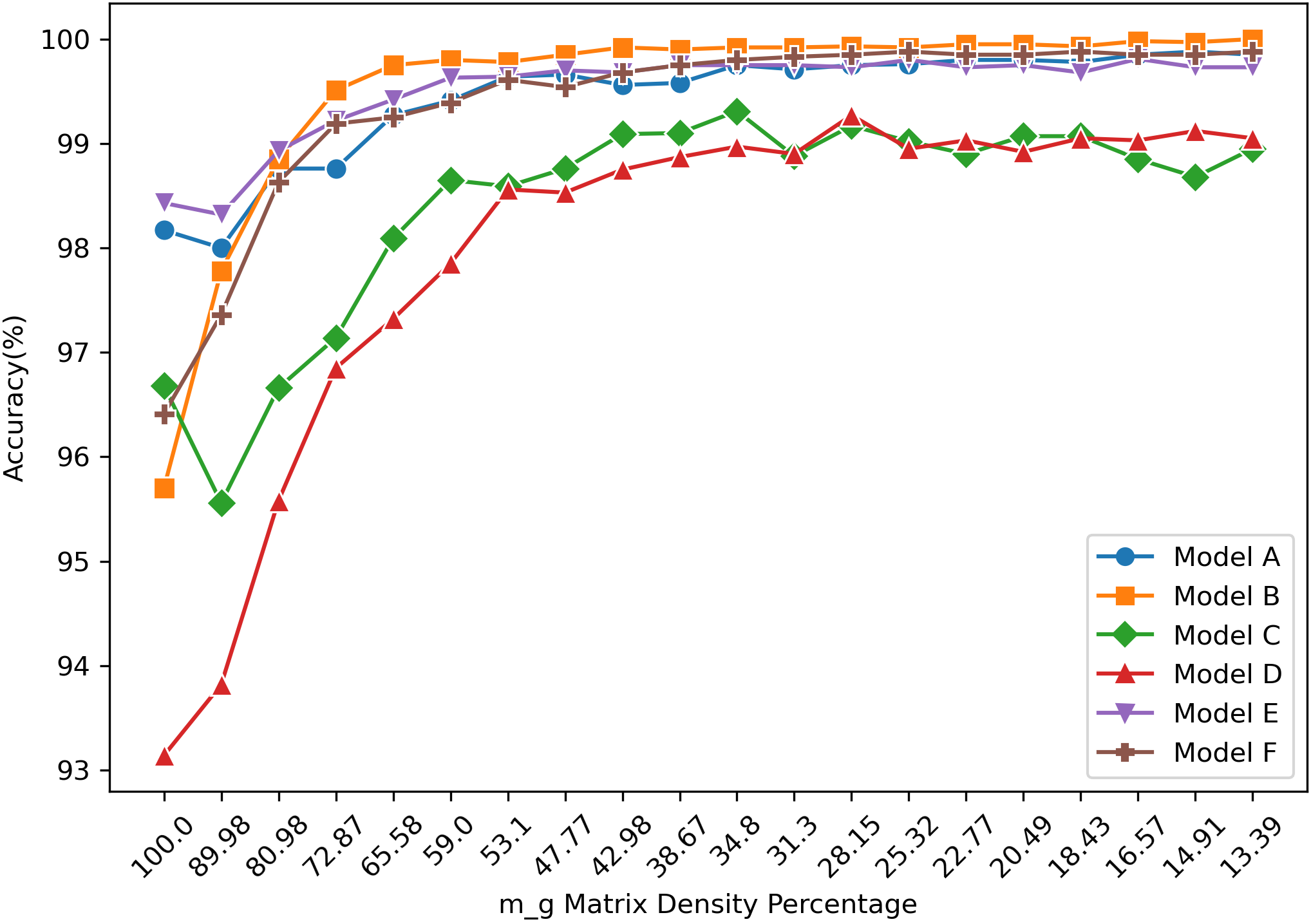}
        \caption{Subject $S_{12}$}
        \label{fig: Subj-12-acc}
    \end{subfigure}
    \hfill
    \begin{subfigure}[t]{0.32\textwidth}
        \centering
        \includegraphics[width=1\textwidth]{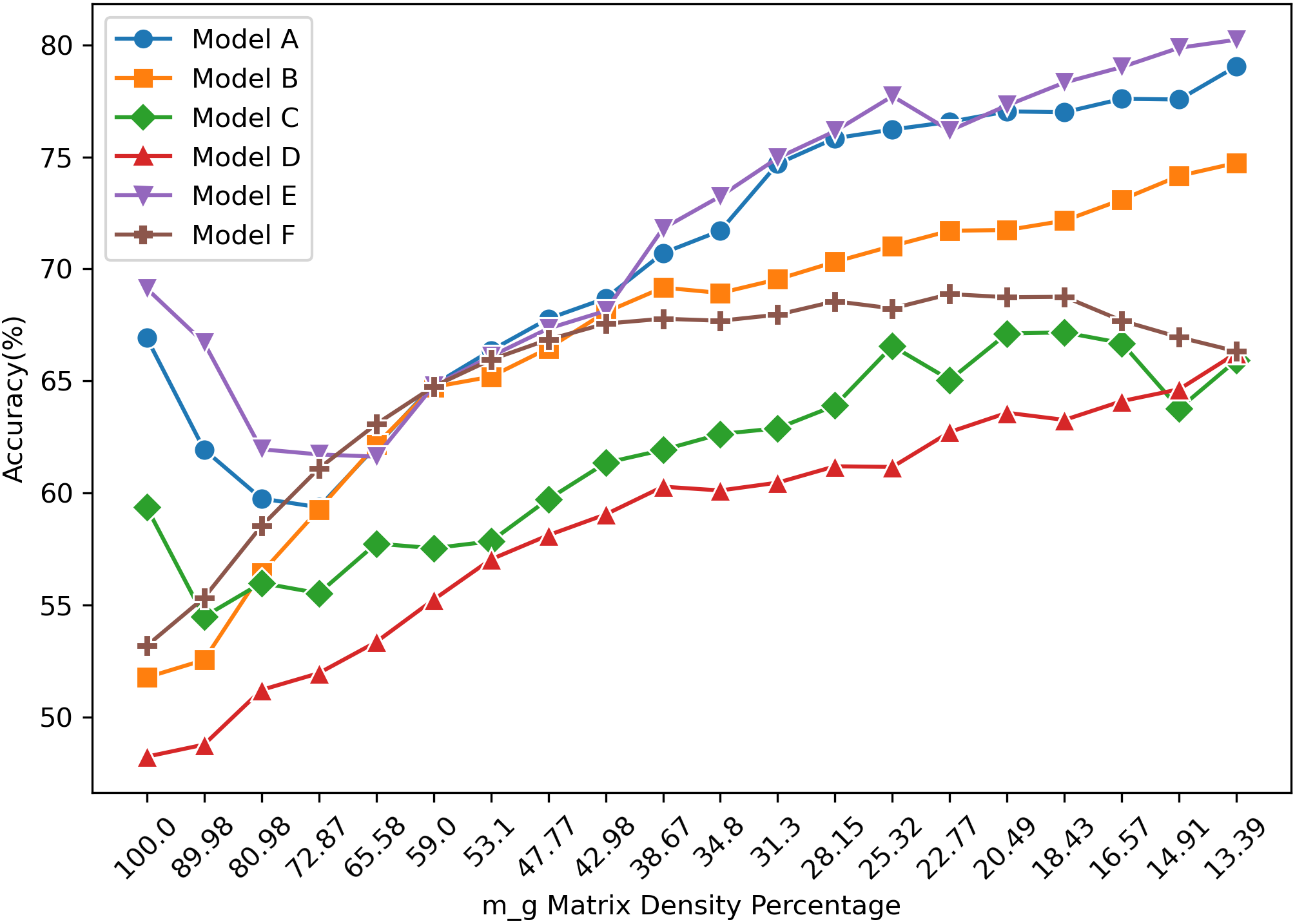}
        \caption{Subject $S_{14}$}
        \label{fig: Subj-14-acc}        
    \end{subfigure}
    \hfill
    \begin{subfigure}[t]{0.32\textwidth}
        \centering
        \includegraphics[width=1\textwidth]{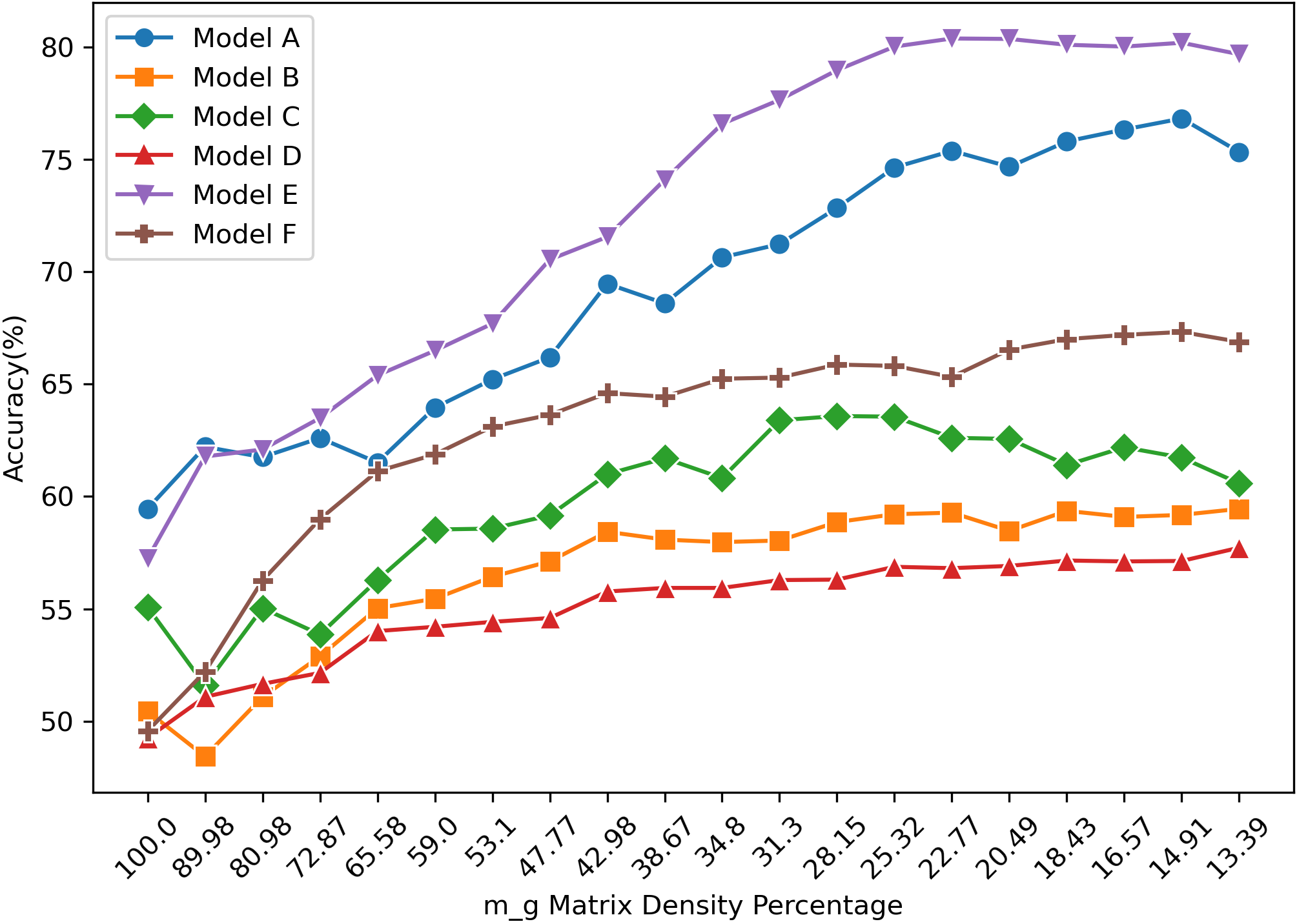}
        \caption{Subject $S_{15}$}
        \label{fig: Subj-15-acc}
    \end{subfigure}
    
    \caption{Performance accuracy across different $m_g$ densities using different models for Subjects $S_1$, $S_3$, $S_6$, $S_{12}$, $S_{14}$ and $S_{15}$ 
    Accuracy vs $m_{g}$ Densities}
    \label{fig: Acc-mg}
\end{figure*}

Figure~\ref{fig: Acc-mg} presents the classification accuracy across various adjacency matrix densities for Subjects $S_1$, $S_3$, $S_6$, $S_{12}$, $S_{14}$ and $S_{15}$. The data indicates an upward trend in classification accuracy with iterative pruning. Most importantly, the accuracy is notably lower at an adjacency matrix density of 100\% in comparison to other densities. This observation suggests that some initial connections between EEG electrodes might be unnecessary, or even counterproductive, for achieving optimal classification. Removing these redundant links may boost the classification accuracy. Hence, a fully connected model between EEG channels may not be the most effective approach.

\begin{figure*}
    \centering
    \includegraphics[width=0.9\textwidth]{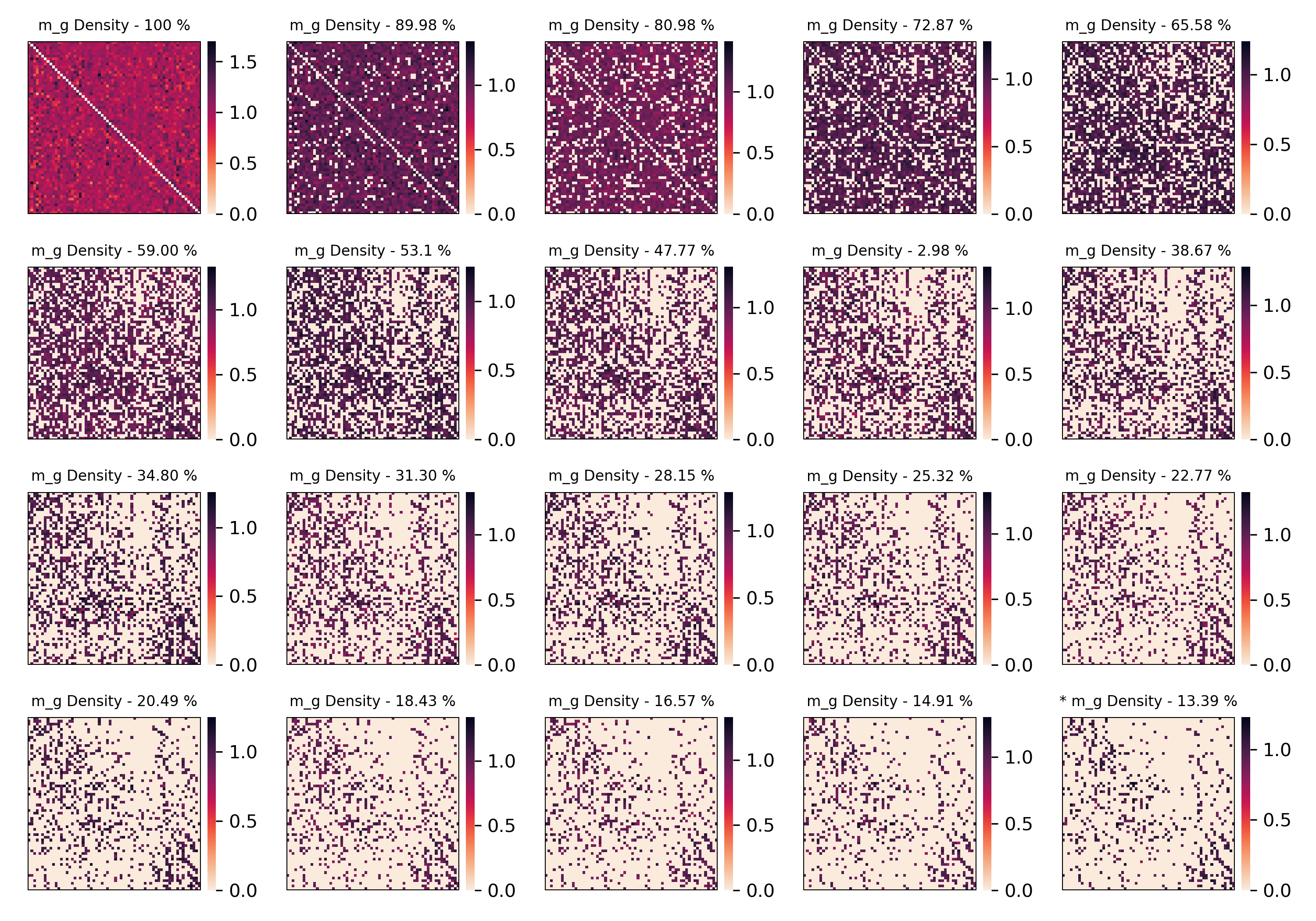}
    \caption{EEG\_GLT Adjacency matrix mask $(m_g)$ of Subject $S_6$ at different densities using Model A. The $m_g$ density at 13.39\% produces the highest accuracy of 78.13\%}
    \label{fig: subj_6-adj-mask-transform}
\end{figure*}

\begin{figure*}
    \centering
    \includegraphics[width=0.9\textwidth]{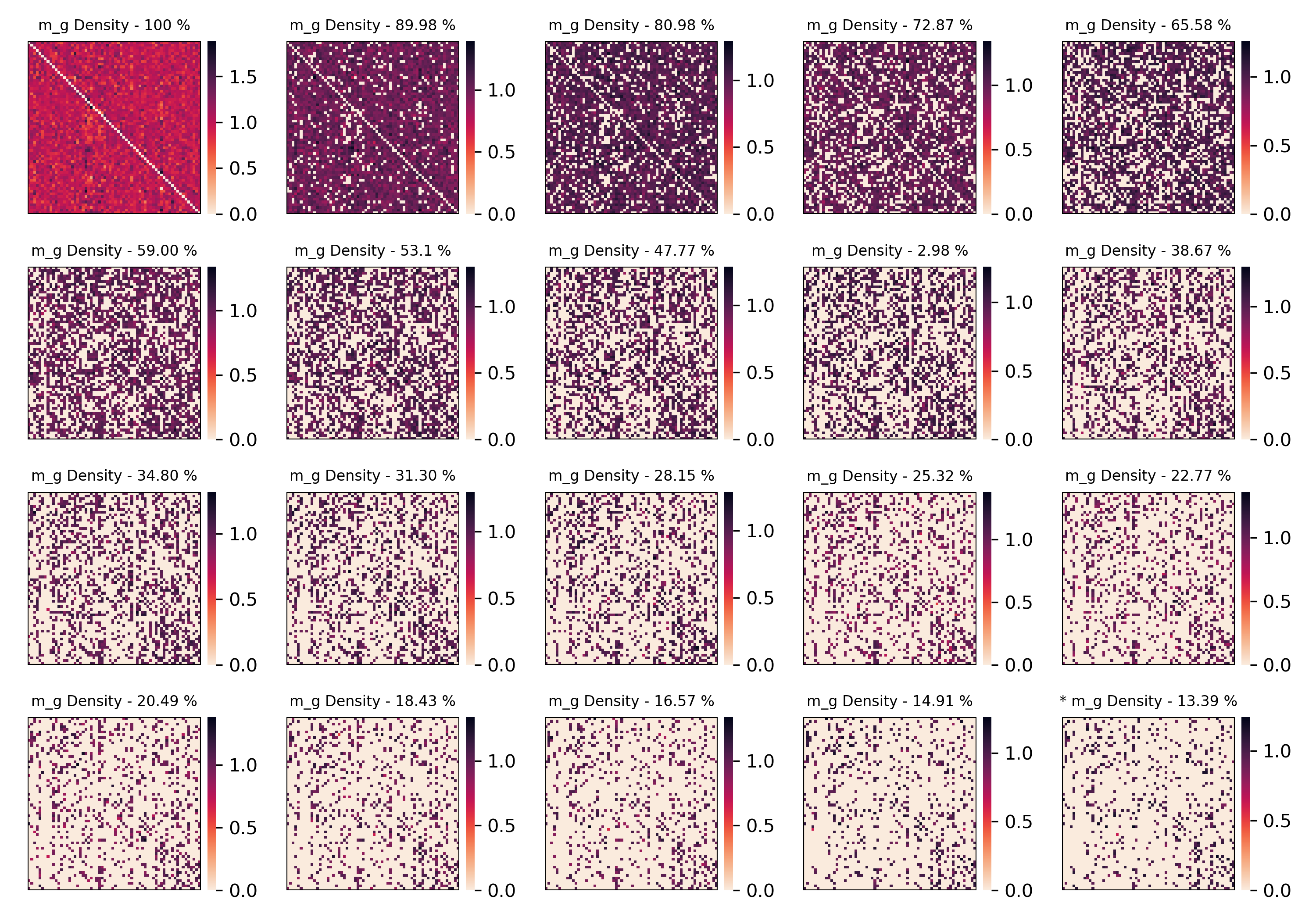}
    \caption{EEG\_GLT Adjacency matrix mask $(m_g)$ of Subject $S_{14}$ at different densities using Model A. The $m_g$ density at 13.39\% produces the highest accuracy of 79.06\%}
    \label{fig: subj_14-adj-mask-transform}
\end{figure*}

Table~\ref{table: Model_accuracy_comparison} displays the optimal EEG\_GLT adjacency matrix ($m_{g\_EEG\_GLT}$) density for each subject. The transformation of the adjacency matrix mask $m_g$ for the subjects $S_6$ and $S_{14}$ at different densities is shown in Figure~\ref{fig: subj_6-adj-mask-transform} and Figure~\ref{fig: subj_14-adj-mask-transform} respectively. For subjects $S_5$, $S_7$, $S_{13}$, and $S_{19}$, their optimal $m_{g\_EEG\_GLT}$ were identified early at a 100\% density. In contrast, other subjects attained their best results at densities below 22.53\% for $2^{nd}$ order models. When considering $5^{th}$ order models, such as Model B, Model D, and Model F, the optimal EEG\_GLTs emerged at densities of 59.00\% or lower.

\begin{table}[ht]
\caption{Optimal EEG\_GLT Adjacency Matrix ($m_{g\_EEG\_GLT}$) Density of Each Subject Across Models}
\label{table: adj density}
\centering
\setlength{\tabcolsep}{1.0pt}
\begin{tabular}{m{1.0cm}<{\centering}  m{1.2cm}<{\centering} m{1.2cm}<{\centering} m{1.2cm}<{\centering}  m{1.2cm}<{\centering}  m{1.2cm}<{\centering}  m{1.2cm}<{\centering}}

\hline
\hline
Subject & Model A & Model B & Model C & Model D & Model E & Model F \\
\hline
$S_1$ & 18.43\% & 13.39\% & 31.30\% & 28.15\% & 18.43\% & 13.39\% \\
$S_2$ & 16.57\% & 13.39\% & 13.39\% & 28.15\% & 18.43\% & 25.32\% \\
$S_3$ & 18.43\% & 25.32\% & 34.80\% & 31.30\% & 25.32\% & 20.49\% \\
$S_4$ & 13.39\% & 13.39\% & 14.91\% & 20.49\% & 14.91\% & 13.39\% \\
$S_5$ & 100.00\% & 31.30\% & 100.00\% & 100.00\% & 100.00\% & 100.00\% \\
$S_6$ & 13.39\% & 20.49\% & 100.00\% & 14.91\% & 14.91\% & 20.49\% \\
$S_7$ & 100.00\% & 28.15\% & 100.00\% & 31.30\% & 100.00\% & 59.00\% \\
$S_8$ & 20.49\% & 18.43\% & 13.39\% & 14.91\% & 31.30\% & 14.91\% \\
$S_9$ & 13.39\% & 16.57\% & 16.57\% & 14.91\% & 13.39\% & 13.39\% \\
$S_{10}$ & 13.39\% & 13.39\% & 22.77\% & 20.49\% & 13.39\% & 13.39\% \\
$S_{11}$ & 13.39\% & 13.39\% & 16.57\% & 13.39\% & 13.39\% & 13.39\% \\
$S_{12}$ & 14.91\% & 13.39\% & 34.80\% & 28.15\% & 16.57\% & 13.39\% \\
$S_{13}$ & 80.98\% & 34.80\% & 100.00\% & 20.49\% & 100.00\% & 22.77\% \\
$S_{14}$ & 13.39\% & 13.39\% & 18.43\% & 13.39\% & 13.39\% & 22.77\% \\
$S_{15}$ & 14.91\% & 13.39\% & 28.15\% & 13.39\% & 22.77\% & 14.91\% \\
$S_{16}$ & 14.91\% & 13.39\% & 20.49\% & 18.43\% & 13.39\% & 13.39\% \\
$S_{17}$ & 14.91\% & 13.39\% & 20.49\% & 22.77\% & 13.39\% & 13.39\% \\
$S_{18}$ & 14.91\% & 13.39\% & 28.15\% & 20.49\% & 22.77\% & 31.30\% \\
$S_{19}$ & 100.00\% & 59.00\% & 100.00\% & 22.77\% & 100.00\% & 31.30\% \\
$S_{20}$ & 25.32\% & 22.77\% & 34.80\% & 16.57\% & 20.49\% & 34.80\% \\

\hline
\hline

\end{tabular}
\end{table}

While our approach enhanced the accuracy for subjects $S_5$, $S_7$, $S_{13}$, and $S_{19}$, the results for both accuracy and F1 score lingered below 60.00\%. A potential explanation is that relying on a single time point feature from EEG channels might not be adequate for MI tasks in these subjects, since there is inherent variability in the time required (or temporal dynamics) to execute the MI task among different individuals, as referenced in \cite{temporal_dynamic}. This variability might also explain why eliminating edges between EEG channels does not necessarily lead to improved performance accuracy for those subjects.

\subsection{Model Setting vs Adjacency Matrix Construction Methods}

Based on Table~\ref{table: Model_accuracy_comparison}, for the Geodesic method, $2^{nd}$ order GCN filters classify with higher average accuracy and F1 score than $5^{th}$ order filters. However, for the PCC and EEG GLT methods, $5^{th}$ order GCN filters perform better. As highlighted in Section~\ref{subsec: eeg_vs_pcc}, our EEG\_GLT method consistently achieves better accuracy than both the PCC and Geodesic methods. This remains the case even when the EEG\_GLT adjacency matrix is paired with Model D, characterised by its minimal complexity, encompassing just five spectral GCN layers with $2^{nd}$ order filters and a singular FC layer. These findings suggest that optimising the adjacency matrix is more importance than refining the GCN architecture when aiming for enhanced performance accuracy.

\subsection{MACs Saving using EEG\_GLT Method}

The MACs inference for classifying a single-time-point EEG MI signal is influenced by several model settings, including the model framework, the number and polynomial order of GCN filters, and the specifications of FC layers as the number of layers and the node count. Among these, the count and polynomial orders of GCN filters at the GCN layers are the primary determinants of the MACs requirement. Both $A^{Geodesic}$ and $A^{PCC}$ maintain 100\% densities in their adjacency matrices. Consequently, the MACs inference for a single-time-point EEG MI signal, when using models A to F, are as follows: 81.89M, 42.26M, 22.64M, 11.32M, 291.62M, and 146.10M, respectively.

\begin{table}[ht]
\caption{MACs Savings (\%) for Each Subject: PCC's Best Model Accuracy vs. EEG\_GLT Accuracy from Models with Adjacency Matrix Densities Just Surpassing PCC's Best Accuracy}
\label{table: Subj_mac_comparison}
\centering
\setlength{\tabcolsep}{3pt}
\begin{tabular}{m{0.6cm}<{\centering} m{0.60cm}<{\centering} m{1.00cm}<{\centering} m{0.80cm}<{\centering}
m{0.15cm}<{\centering}    m{1.35cm}<{\centering} m{0.90cm}<{\centering} m{0.90cm}<{\centering} 
 m{0.001cm}<{\centering}   m{0.70cm}<{\centering}}
\hline
\hline
\multirow{2}{*}{Subj} & \multicolumn{3}{c}{PCC} & & \multicolumn{3}{c}{EEG\_GLT} & & MACs\\
\cline{2-4} \cline{6-8}
 & Model & Acc. & MACs & & Model \newline(Adj\%) & Acc. & MACs & & Saving \\
\hline
$S_{1}$ & A & 87.66\% & 81.89M
        & & D (13.39\%)  & 97.04\% & \textbf{8.76M}
        & & 
        89.30\%\\
$S_{2}$ & B & 75.43\% & 42.26M
        & & B (13.39\%)  & 78.09\% & \textbf{36.97M}
        & & 
        12.52\%\\
$S_{3}$ & A & 94.89\% & 81.89M
        & & D (13.39\%)  & 98.22\% & \textbf{8.76M}
        & & 
        89.30\%\\
$S_{4}$ & A & 99.88\% & 81.89M
        & & B (13.39\%)  & 99.98\% & \textbf{36.97M}
        & & 
        54.85\%\\
$S_{5}$ & B & 46.90\% & 42.26M
        & & B (13.39\%)  & 48.73\% & \textbf{36.97M}
        & & 
        12.52\%\\
$S_{6}$ & E & 62.92\% & 291.62M
        & & B (13.39\%)  & 70.17\% & \textbf{36.97M}
        & & 
        87.32\%\\
$S_{7}$ & E & 55.04\% & 291.62M
        & & B (13.39\%)  & 57.68\% & \textbf{36.97M}
        & & 
        87.32\%\\
$S_{8}$ & B & 98.71\% & 42.26M
        & & D (13.39\%)  & 99.78\% & \textbf{8.76M}
        & & 
        79.27\%\\
$S_{9}$ & A & 99.86\% & 81.89M
        & & B (13.39\%)  & 99.98\% & \textbf{36.97M}
        & & 
        54.85\%\\
$S_{10}$ & E & 99.44\% & 291.62M
        & & D (13.39\%)  & 99.97\% & \textbf{8.76M}
        & & 
        97.00\%\\
$S_{11}$ & E & 99.90\% & 291.62M
        & & D (13.39\%)  & 99.98\% & \textbf{8.76M}
        & & 
        97.00\%\\
$S_{12}$ & A & 86.76\% & 81.89M
        & & D (13.39\%)  & 99.05\% & \textbf{8.76M}
        & & 
        89.30\%\\
$S_{13}$ & A & 42.79\% & 81.89M
        & & B (13.39\%)  & 43.57\% & \textbf{36.97M}
        & & 
        54.85\%\\
$S_{14}$ & B & 63.58\% & 42.26M
        & & D (13.39\%)  & 66.25\% & \textbf{8.76M}
        & & 
        79.29\%\\
$S_{15}$ & E & 57.01\% & 291.62M
        & & D (13.39\%)  & 57.72\% & \textbf{8.76M}
        & & 
        97.00\%\\
$S_{16}$ & B & 99.80\% & 42.26M
        & & D (13.39\%)  & 99.85\% & \textbf{8.76M}
        & & 
        79.27\%\\
$S_{17}$ & A & 99.98\% & 81.89M
        & & B (13.39\%)  & 100.00\% & \textbf{36.97M}
        & & 
        44.93\%\\
$S_{18}$ & A & 96.05\% & 81.89M
        & & D (16.57\%)  & 99.58\% & \textbf{8.76M}
        & & 
        76.14\%\\
$S_{19}$ & A & 41.62\% & 81.89M
        & & A (89.98\%)  & 41.78\% & \textbf{80.67M}
        & & 
        1.49\%\\
$S_{20}$ & B & 99.17\% & 42.26M
        & & D (13.39\%)  & 99.68\% & \textbf{8.76M}
        & & 
        79.27\%\\

\hline
\hline
\end{tabular}
\end{table}

Our EEG\_GLT method presents varied $A^{EEG\_GLT}$ densities due to the pruning employed by Algorithm~\ref{alg: EEG_GLT}. As elaborated in Section~\ref{subsec: eeg_vs_pcc}, the EEG\_GLT approach enhances classification accuracy through pruning, which in turn decreases the MACs. Table~\ref{table: Subj_mac_comparison} illustrates the percentage of MACs savings for each subject, comparing the top accuracy value from the PCC method to the EEG\_GLT accuracies from models with adjacency matrix densities slightly exceeding PCC's best.

For performance equivalent to or surpassing PCC's optimal accuracy, only Models D and B with the sparsest adjacency matrix density (13.39\%) are necessary. The PCC method requires between 42.26M to 291.62M for one-time-point inference across 20 subjects to reach peak accuracy. In contrast, our EEG\_GLT approach needs only 8.76M to 80.67M to achieve equal or better accuracy, translating to savings in MACs of up to 97.00\%.

\section{Conclusion}

Our EEG\_GLT approach, developed for optimal adjacency matrix construction in EEG MI time-point signal classification, consistently outperforms both the Geodesic and PCC methods in accuracy and F1 score. It is important to  note that the PCC method is currently employed in the state-of-the-art EEG time-point classification model, GCNs-Net. Specifically, our EEG\_GLT method enhances accuracy and F1 score by margins ranging from 0.52\% to 22.04\% and 0.50\% to 21.76\%, respectively, compared to PCC. Furthermore, it improves the average accuracy across 20 subjects by 13.39\%. With this method, optimal outcomes emerge when the adjacency matrix densities remain below 22.53\%. Our study emphasises the pivotal role played by the configuration of the adjacency matrix in performance accuracy, overshadowing even model settings. In addition, our EEG\_GLT approach has much higher computational efficiency, demanding between 8.76M and 80.67M MACs, which is significantly less than the 42.26M to 291.62M required by the PCC method for comparable or superior results.

While this research primarily focuses on identifying the optimal adjacency matrix, with pruning confined to the adjacency matrix, upcoming studies will explore pruning GNN and FC layers weights to further streamline computational costs. Additionally, we plan to expand the number of time points used for feature extraction, especially for subjects $S_5$, $S_7$, $S_{13}$, and $S_{19}$. In future work, we will refine Algorithm~\ref{alg: EEG_GLT} to seamlessly integrate pooling layers within the GCN blocks under the EEG\_GLT method, to further optimise computational efficiency. Our approach effectively constructs the adjacency matrix to capture the optimal relationship between EEG channels, but it primarily targets EEG MI tasks. To achieve a more generalised understanding of the inter-relationships between EEG channels, it is essential to incorporate a broader range of tasks into models.

\section{Acknowledgement}
The author, Htoo Wai Aung, extends sincere gratitude to Prof. Steven Su, Dr. Jiao Jiao Li, and Dr. Yang An for their invaluable guidance and support throughout this research. Special thanks are due to the University of Technology Sydney, the School of Biomedical Engineering, and the UTS Research Excellence Scholarship for their steadfast support. Additionally, heartfelt appreciation is extended to my family, whose contributions were essential to the completion of this article.



\begin{thebibliography}{10}
\providecommand{\url}[1]{#1}
\csname url@samestyle\endcsname
\providecommand{\newblock}{\relax}
\providecommand{\bibinfo}[2]{#2}
\providecommand{\BIBentrySTDinterwordspacing}{\spaceskip=0pt\relax}
\providecommand{\BIBentryALTinterwordstretchfactor}{4}
\providecommand{\BIBentryALTinterwordspacing}{\spaceskip=\fontdimen2\font plus
\BIBentryALTinterwordstretchfactor\fontdimen3\font minus \fontdimen4\font\relax}
\providecommand{\BIBforeignlanguage}[2]{{%
\expandafter\ifx\csname l@#1\endcsname\relax
\typeout{** WARNING: IEEEtran.bst: No hyphenation pattern has been}%
\typeout{** loaded for the language `#1'. Using the pattern for}%
\typeout{** the default language instead.}%
\else
\language=\csname l@#1\endcsname
\fi
#2}}
\providecommand{\BIBdecl}{\relax}
\BIBdecl

\bibitem{WOLPAW2002767}
\BIBentryALTinterwordspacing
J.~R. Wolpaw, N.~Birbaumer, D.~J. McFarland, G.~Pfurtscheller, and T.~M. Vaughan, ``Brain–computer interfaces for communication and control,'' \emph{Clinical Neurophysiology}, vol. 113, no.~6, pp. 767--791, 2002. [Online]. Available: \url{https://www.sciencedirect.com/science/article/pii/S1388245702000573}
\BIBentrySTDinterwordspacing

\bibitem{lebedev2006brain}
M.~A. Lebedev and M.~A. Nicolelis, ``Brain--machine interfaces: past, present and future,'' \emph{TRENDS in Neurosciences}, vol.~29, no.~9, pp. 536--546, 2006.

\bibitem{schomer2012niedermeyer}
D.~L. Schomer and F.~L. Da~Silva, \emph{Niedermeyer's electroencephalography: basic principles, clinical applications, and related fields}.\hskip 1em plus 0.5em minus 0.4em\relax Lippincott Williams \& Wilkins, 2012.

\bibitem{hubbard2019eeg}
J.~Hubbard, A.~Kikumoto, and U.~Mayr, ``Eeg decoding reveals the strength and temporal dynamics of goal-relevant representations,'' \emph{Scientific reports}, vol.~9, no.~1, p. 9051, 2019.

\bibitem{mcfarland2000mu}
D.~J. McFarland, L.~A. Miner, T.~M. Vaughan, and J.~R. Wolpaw, ``Mu and beta rhythm topographies during motor imagery and actual movements,'' \emph{Brain topography}, vol.~12, pp. 177--186, 2000.

\bibitem{jeannerod1994representing}
M.~Jeannerod, ``The representing brain: Neural correlates of motor intention and imagery,'' \emph{Behavioral and Brain sciences}, vol.~17, no.~2, pp. 187--202, 1994.

\bibitem{biasiucci2018brain}
A.~Biasiucci, R.~Leeb, I.~Iturrate, S.~Perdikis, A.~Al-Khodairy, T.~Corbet, A.~Schnider, T.~Schmidlin, H.~Zhang, M.~Bassolino \emph{et~al.}, ``Brain-actuated functional electrical stimulation elicits lasting arm motor recovery after stroke,'' \emph{Nature communications}, vol.~9, no.~1, p. 2421, 2018.

\bibitem{farabet2012learning}
C.~Farabet, C.~Couprie, L.~Najman, and Y.~LeCun, ``Learning hierarchical features for scene labeling,'' \emph{IEEE transactions on pattern analysis and machine intelligence}, vol.~35, no.~8, pp. 1915--1929, 2012.

\bibitem{lecun1998gradient}
Y.~LeCun, L.~Bottou, Y.~Bengio, and P.~Haffner, ``Gradient-based learning applied to document recognition,'' \emph{Proceedings of the IEEE}, vol.~86, no.~11, pp. 2278--2324, 1998.

\bibitem{lecun2010convolutional}
Y.~LeCun, K.~Kavukcuoglu, and C.~Farabet, ``Convolutional networks and applications in vision,'' in \emph{Proceedings of 2010 IEEE international symposium on circuits and systems}.\hskip 1em plus 0.5em minus 0.4em\relax IEEE, 2010, pp. 253--256.

\bibitem{zhang2020deep}
Z.~Zhang, P.~Cui, and W.~Zhu, ``Deep learning on graphs: A survey,'' \emph{IEEE Transactions on Knowledge and Data Engineering}, vol.~34, no.~1, pp. 249--270, 2020.

\bibitem{wu2020comprehensive}
Z.~Wu, S.~Pan, F.~Chen, G.~Long, C.~Zhang, and S.~Y. Philip, ``A comprehensive survey on graph neural networks,'' \emph{IEEE transactions on neural networks and learning systems}, vol.~32, no.~1, pp. 4--24, 2020.

\bibitem{bruna2013spectral}
J.~Bruna, W.~Zaremba, A.~Szlam, and Y.~LeCun, ``Spectral networks and locally connected networks on graphs,'' \emph{arXiv preprint arXiv:1312.6203}, 2013.

\bibitem{defferrard2016convolutional}
M.~Defferrard, X.~Bresson, and P.~Vandergheynst, ``Convolutional neural networks on graphs with fast localized spectral filtering,'' \emph{Advances in neural information processing systems}, vol.~29, 2016.

\bibitem{levie2018cayleynets}
R.~Levie, F.~Monti, X.~Bresson, and M.~M. Bronstein, ``Cayleynets: Graph convolutional neural networks with complex rational spectral filters,'' \emph{IEEE Transactions on Signal Processing}, vol.~67, no.~1, pp. 97--109, 2018.

\bibitem{hamilton2017inductive}
W.~Hamilton, Z.~Ying, and J.~Leskovec, ``Inductive representation learning on large graphs,'' \emph{Advances in neural information processing systems}, vol.~30, 2017.

\bibitem{monti2017geometric}
F.~Monti, D.~Boscaini, J.~Masci, E.~Rodola, J.~Svoboda, and M.~M. Bronstein, ``Geometric deep learning on graphs and manifolds using mixture model cnns,'' in \emph{Proceedings of the IEEE conference on computer vision and pattern recognition}, 2017, pp. 5115--5124.

\bibitem{niepert2016learning}
M.~Niepert, M.~Ahmed, and K.~Kutzkov, ``Learning convolutional neural networks for graphs,'' in \emph{International conference on machine learning}.\hskip 1em plus 0.5em minus 0.4em\relax PMLR, 2016, pp. 2014--2023.

\bibitem{gao2018large}
H.~Gao, Z.~Wang, and S.~Ji, ``Large-scale learnable graph convolutional networks,'' in \emph{Proceedings of the 24th ACM SIGKDD international conference on knowledge discovery \& data mining}, 2018, pp. 1416--1424.

\bibitem{bao2022linking}
G.~Bao, K.~Yang, L.~Tong, J.~Shu, R.~Zhang, L.~Wang, B.~Yan, and Y.~Zeng, ``Linking multi-layer dynamical gcn with style-based recalibration cnn for eeg-based emotion recognition,'' \emph{Frontiers in Neurorobotics}, vol.~16, p. 834952, 2022.

\bibitem{shuman2013emerging}
D.~I. Shuman, S.~K. Narang, P.~Frossard, A.~Ortega, and P.~Vandergheynst, ``The emerging field of signal processing on graphs: Extending high-dimensional data analysis to networks and other irregular domains,'' \emph{IEEE signal processing magazine}, vol.~30, no.~3, pp. 83--98, 2013.

\bibitem{zeng2020hierarchy}
D.~Zeng, K.~Huang, C.~Xu, H.~Shen, and Z.~Chen, ``Hierarchy graph convolution network and tree classification for epileptic detection on electroencephalography signals,'' \emph{IEEE transactions on cognitive and developmental systems}, vol.~13, no.~4, pp. 955--968, 2020.

\bibitem{meng2022electrical}
L.~Meng, J.~Hu, Y.~Deng, and Y.~Hu, ``Electrical status epilepticus during sleep electroencephalogram waveform identification and analysis based on a graph convolutional neural network,'' \emph{Biomedical Signal Processing and Control}, vol.~77, p. 103788, 2022.

\bibitem{hou2022gcns}
Y.~Hou, S.~Jia, X.~Lun, Z.~Hao, Y.~Shi, Y.~Li, R.~Zeng, and J.~Lv, ``Gcns-net: a graph convolutional neural network approach for decoding time-resolved eeg motor imagery signals,'' \emph{IEEE Transactions on Neural Networks and Learning Systems}, 2022.

\bibitem{zhang2022recognizing}
R.~Zhang, Z.~Wang, F.~Yang, and Y.~Liu, ``Recognizing the level of organizational commitment based on deep learning methods and eeg,'' in \emph{ITM Web of Conferences}, vol.~47.\hskip 1em plus 0.5em minus 0.4em\relax EDP Sciences, 2022, p. 02044.

\bibitem{jia2022efficient}
M.~Jia, W.~Liu, J.~Duan, L.~Chen, C.~Chen, Q.~Wang, and Z.~Zhou, ``Efficient graph convolutional networks for seizure prediction using scalp eeg,'' \emph{Frontiers in Neuroscience}, vol.~16, p. 967116, 2022.

\bibitem{wagh2020eeg}
N.~Wagh and Y.~Varatharajah, ``Eeg-gcnn: Augmenting electroencephalogram-based neurological disease diagnosis using a domain-guided graph convolutional neural network,'' in \emph{Machine Learning for Health}.\hskip 1em plus 0.5em minus 0.4em\relax PMLR, 2020, pp. 367--378.

\bibitem{ma2023double}
W.~Ma, C.~Wang, X.~Sun, X.~Lin, and Y.~Wang, ``A double-branch graph convolutional network based on individual differences weakening for motor imagery eeg classification,'' \emph{Biomedical Signal Processing and Control}, vol.~84, p. 104684, 2023.

\bibitem{khaleghi2023developing}
N.~Khaleghi, T.~Y. Rezaii, S.~Beheshti, and S.~Meshgini, ``Developing an efficient functional connectivity-based geometric deep network for automatic eeg-based visual decoding,'' \emph{Biomedical Signal Processing and Control}, vol.~80, p. 104221, 2023.

\bibitem{song2018eeg}
T.~Song, W.~Zheng, P.~Song, and Z.~Cui, ``Eeg emotion recognition using dynamical graph convolutional neural networks,'' \emph{IEEE Transactions on Affective Computing}, vol.~11, no.~3, pp. 532--541, 2018.

\bibitem{chen2021unified}
T.~Chen, Y.~Sui, X.~Chen, A.~Zhang, and Z.~Wang, ``A unified lottery ticket hypothesis for graph neural networks,'' in \emph{International conference on machine learning}.\hskip 1em plus 0.5em minus 0.4em\relax PMLR, 2021, pp. 1695--1706.

\bibitem{physionet_dataset}
A.~L. Goldberger, L.~A. Amaral, L.~Glass, J.~M. Hausdorff, P.~C. Ivanov, R.~G. Mark, J.~E. Mietus, G.~B. Moody, C.-K. Peng, and H.~E. Stanley, ``Physiobank, physiotoolkit, and physionet: components of a new research resource for complex physiologic signals,'' \emph{circulation}, vol. 101, no.~23, pp. e215--e220, 2000.

\bibitem{van1998volume}
S.~P. van~den Broek, F.~Reinders, M.~Donderwinkel, and M.~Peters, ``Volume conduction effects in eeg and meg,'' \emph{Electroencephalography and clinical neurophysiology}, vol. 106, no.~6, pp. 522--534, 1998.

\bibitem{temporal_dynamic}
C.~Tallon-Baudry and O.~Bertrand, ``Oscillatory gamma activity in humans and its role in object representation,'' \emph{Trends in cognitive sciences}, vol.~3, no.~4, pp. 151--162, 1999.

\end{thebibliography}
\end{document}